\documentclass[letterpaper]{article} % DO NOT CHANGE THIS
\usepackage{aaai2027} % Camera-ready AAAI style
\usepackage[hyphens]{url} % DO NOT CHANGE THIS
\usepackage{graphicx} % DO NOT CHANGE THIS
\usepackage{natbib} % DO NOT CHANGE THIS AND DO NOT ADD ANY OPTIONS TO IT
\usepackage{caption} % DO NOT CHANGE THIS AND DO NOT ADD ANY OPTIONS TO IT
\usepackage{amsmath}
\usepackage{amssymb}
\usepackage{booktabs}
\usepackage{multirow}
\usepackage{tabularx}
\usepackage[table]{xcolor}

\definecolor{bestcolor}{RGB}{245,190,190}    % best: red
\definecolor{secondcolor}{RGB}{244,214,176}  % second: orange
\definecolor{thirdcolor}{RGB}{245,239,186}   % third: yellow

\newcolumntype{Y}{>{\centering\arraybackslash}X}

\title{I3DM: Implicit 3D-aware Memory Retrieval and Injection \\ for Consistent Video Scene Generation}
\author{
Jia Li\textsuperscript{\rm 1,2}\thanks{Work done during internship at Vertex Lab.},
Han Yan\textsuperscript{\rm 2,3},
Yihang Chen\textsuperscript{\rm 1,3},
Siqi Li\textsuperscript{\rm 2},\\
Xibin Song\textsuperscript{\rm 2},
Yifu Wang\textsuperscript{\rm 2},
Jianfei Cai\textsuperscript{\rm 1},
Tien-Tsin Wong\textsuperscript{\rm 1},
Pan Ji\textsuperscript{\rm 2}
}
\affiliations{
\textsuperscript{\rm 1}Monash University \quad
\textsuperscript{\rm 2}Vertex Lab \quad
\textsuperscript{\rm 3}Shanghai Jiao Tong University
}

\nocopyright

\makeatletter
\let\aaaioriginalmaketitle\@maketitle
\def\@maketitle{%
  \aaaioriginalmaketitle
  \begin{minipage}{\textwidth}
    \centering
    \includegraphics[width=0.95\textwidth]{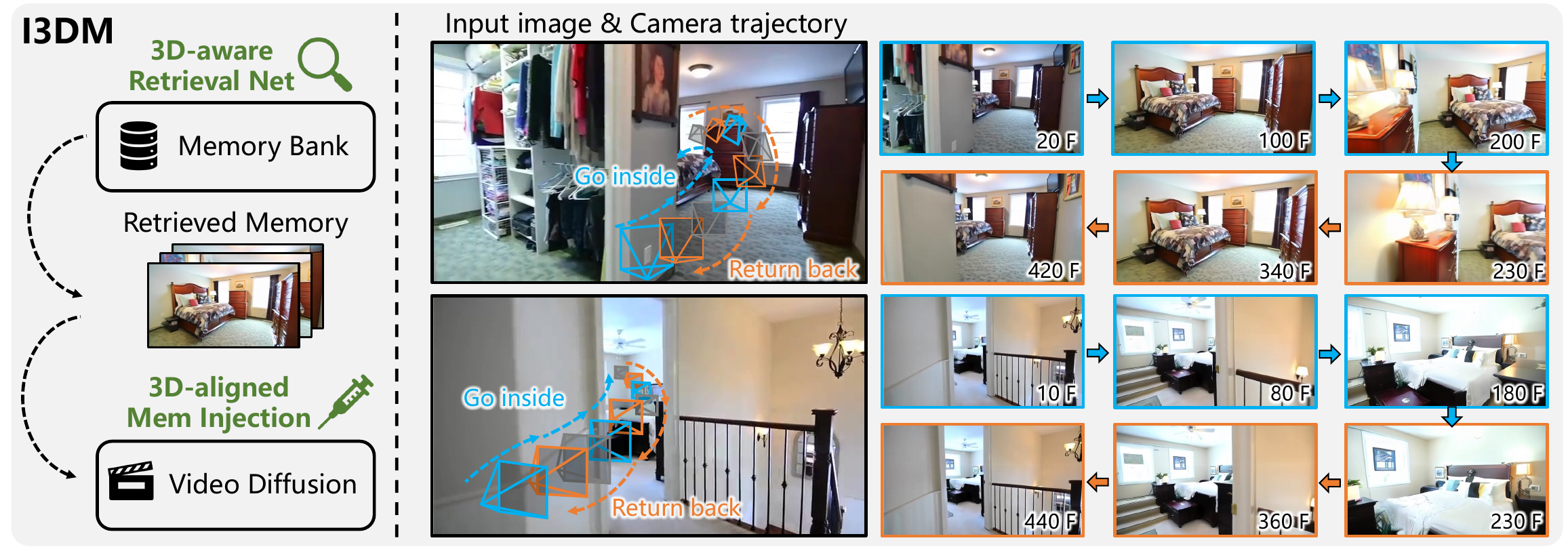}
    \captionof{figure}{Overview of \textbf{I3DM}, an implicit 3D-aware memory mechanism for consistent video generation.
    Given an input image and a user-specified camera trajectory, I3DM enables consistent scene exploration via a 3D-aware memory retrieval network and a 3D-aligned memory injection module. Our method ensures consistent revisiting even under complex occlusions; e.g., after entering an occluded room (blue trajectory), the camera can return to the same area observed earlier (orange trajectory).}
    \vspace{+2em}
    \label{fig:teaser}
  \end{minipage}
}
\makeatother

\begin{document}

\maketitle

\def\teaseratfront{}

\begin{abstract}
Despite remarkable progress in video generation, maintaining long-term scene consistency upon revisiting previously explored areas remains challenging. Existing solutions rely either on explicitly constructing 3D geometry, which suffers from error accumulation and scale ambiguity, or on naive camera Field-of-View (FoV) retrieval, which typically fails under complex occlusions. To overcome these limitations, we propose \textit{I3DM}, a novel implicit 3D-aware memory mechanism for consistent video scene generation that bypasses explicit 3D reconstruction. At the core of our approach is a 3D-aware memory retrieval strategy, which leverages the intermediate features of a pre-trained Feed-Forward Novel View Synthesis (FF-NVS) model to score view relevance, enabling robust retrieval even in highly occluded scenarios. Furthermore, to fully utilize the retrieved historical frames, we introduce a 3D-aligned memory injection module. This module implicitly warps historical content to the target view and adaptively conditions the generation on reliable warping regions, leading to improved revisit consistency and accurate camera control. Extensive experiments demonstrate that our method outperforms state-of-the-art approaches, achieving superior revisit consistency, generation fidelity, and camera control precision. Project page: \url{https://riga2.github.io/i3dm/}.
% \keywords{Consistent Video Generation \and Novel View Synthesis \and Long-term Memory Mechanism}
\end{abstract}

% The original paper uses full-text-width floats. Map them to AAAI's
% two-column float environments without changing the shared section files.
% \renewenvironment{figure}[1][tbp]{\begin{figure*}[#1]}{\end{figure*}}
% \renewenvironment{table}[1][tbp]{\begin{table*}[#1]}{\end{table*}}

% Negative manual spacing from the source paper is not permitted by AAAI.
\let\sourcevspace\vspace
\renewcommand{\vspace}[1]{}

\section{Introduction} \label{sec:intro}

Recent advances in video generation~\cite{qin2024worldsimbench, wan2025wan, brooks2024video, kong2024hunyuanvideo, mao2025yume, parker2024genie} have enabled exploration of diverse and high-fidelity virtual worlds. Given an initial observation and a desired camera trajectory, these models synthesize continuous video streams. 
However, maintaining long-term scene consistency remains challenging due to the absence of long-term memory. As a result, models often exhibit a ``turn-and-forget'' phenomenon: hallucinating inconsistent content upon revisiting previously explored areas, thereby eroding visual realism and plausibility.

% \begin{figure}[h]
%   \centering
%   \includegraphics[width=1.0\textwidth]{fig/intro_v4.pdf}
%   \caption{Limitations of existing memory mechanisms. (Top-Left) Explicit geometry-based memory methods suffer from scale ambiguity during reconstruction, leading to inaccurate camera navigation (e.g., camera collisions) and severe revisit inconsistencies. (Top-Right) Implicit FoV-based memory methods fail in occlusion scenarios, as FoV overlap does not guarantee actual visual visibility. This retrieves irrelevant historical frames, leading to repeated semantic content and inconsistencies during revisits.}
%   \label{fig:intro}
% \end{figure}

% \begin{figure}[t]
%   \centering
%   \includegraphics[width=1.0\textwidth]{fig/teaser_v7.pdf}
%   \caption{Overview of \textbf{I3DM}, an implicit 3D-aware memory mechanism for consistent video generation.
%   Given an input image and a user-specified camera trajectory, I3DM enables consistent scene exploration and region revisiting (indicated by matching colored frames) via a 3D-aware memory retrieval network and a 3D-aligned memory injection module, even for complex occlusion.
%   }
%   \label{fig:teaser}
% \vspace{-1em}
% \end{figure}

\ifdefined\teaseratfront\else
\begin{figure}[t]
  \centering
  \includegraphics[width=1.0\textwidth]{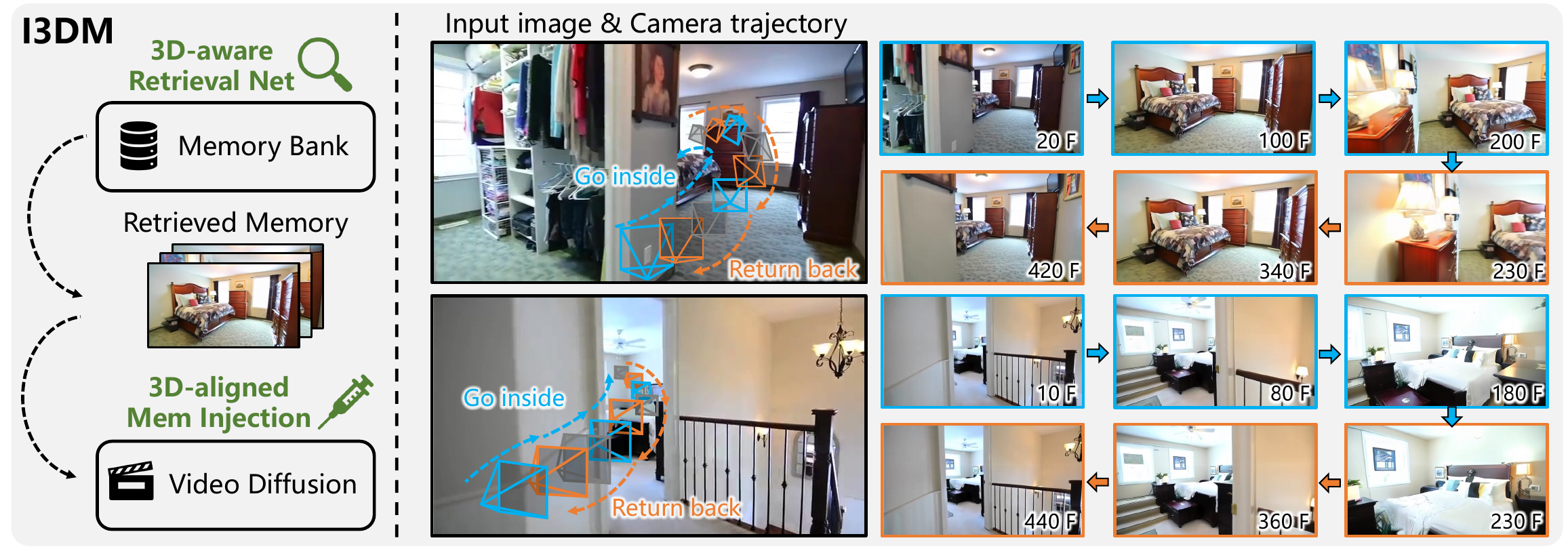}
  \caption{Overview of \textbf{I3DM}, an implicit 3D-aware memory mechanism for consistent video generation.
  Given an input image and a user-specified camera trajectory, I3DM enables consistent scene exploration via a 3D-aware memory retrieval network and a 3D-aligned memory injection module. Our method ensures consistent revisiting (indicated by frames with matching colors), even under complex occlusions.
  }
  \label{fig:teaser}
% \vspace{-0.5em}
\end{figure}
\fi

% \begin{figure*}[h]
%   \centering
%   \includegraphics[width=0.9\linewidth]{fig/intro_new_v2.pdf}
%   \caption{Limitations of existing memory mechanisms. Left: Explicit geometry-based methods (e.g., Gen3C~\cite{ren2025gen3c}) suffer from scale ambiguity, causing inaccurate camera navigation (e.g., colliding with the wall) and revisit inconsistencies. Right: Implicit FoV-based methods (e.g., WorldPlay~\cite{sun2025worldplay}) fail under occlusions, as FoV overlap ignores actual visibility. Retrieving irrelevant historical frames causes repeated content and inconsistent revisits. Frames with matching colors denote the same viewpoint and should be strictly consistent. Red circles and boxes indicate inconsistencies and repeated content.}
%   \label{fig:intro}
% \vspace{-0.5em}
% \end{figure*}

\begin{figure}[h]
  \centering
  \includegraphics[width=\linewidth]{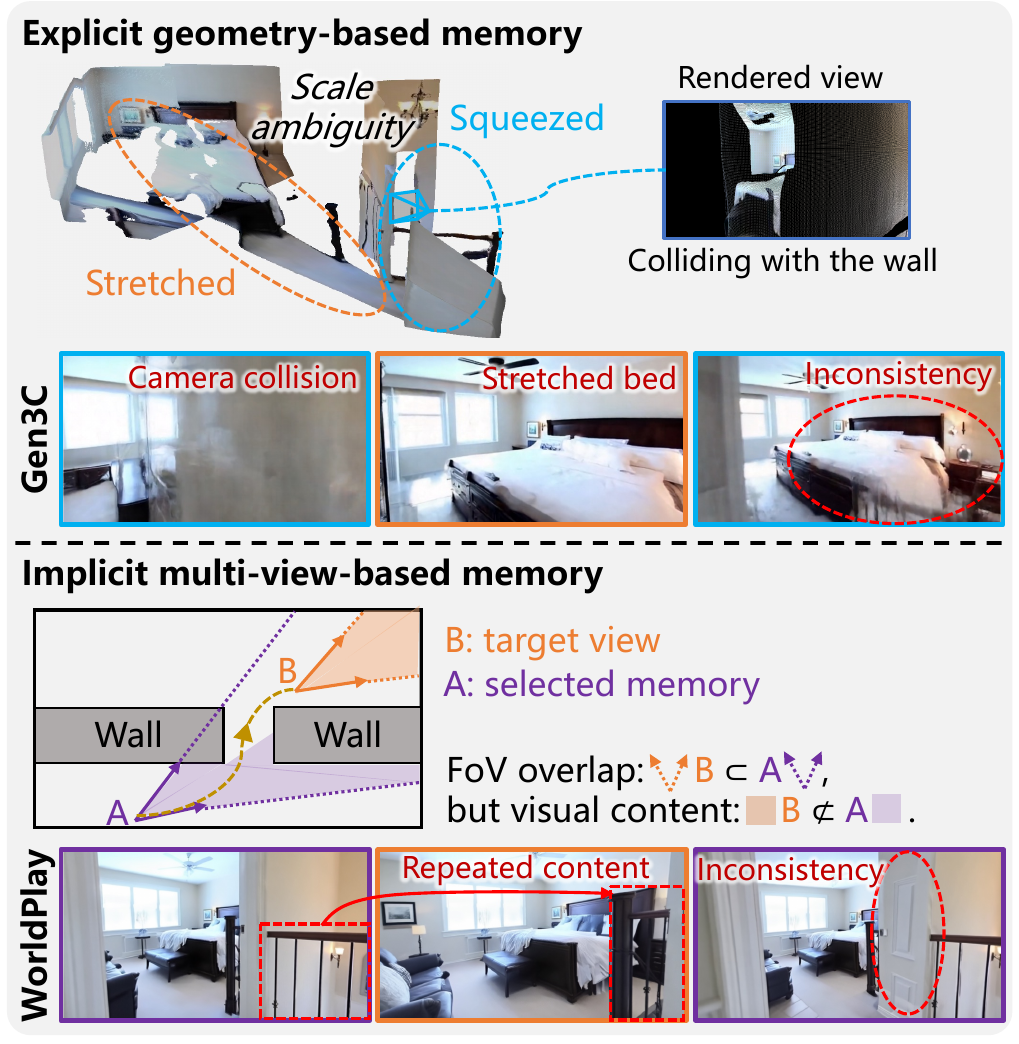}
  \caption{Limitations of existing memory mechanisms. Top: Explicit geometry-based methods (e.g., Gen3C~\cite{ren2025gen3c}) suffer from scale ambiguity, causing inaccurate camera navigation (e.g., colliding with the wall) and revisit inconsistencies. Bottom: Implicit FoV-based methods (e.g., WorldPlay~\cite{sun2025worldplay}) fail under occlusions, as FoV overlap ignores actual visibility. Retrieving irrelevant historical frames causes repeated content and inconsistent revisits. Frames with matching colors denote the same viewpoint and should be strictly consistent. Red circles and boxes indicate inconsistencies and repeated content.}
  \label{fig:intro}
% \vspace{-0.5em}
\end{figure}

% These methods typically utilize feed-forward reconstruction models~\cite{wang2025vggt, wang2025cut3r, chen2025ttt3r, wang2025moge} to estimate 3D geometry from generated views, render partial content for the target view, and employ video generation models to inpaint missing regions. The generated results are then used to  update the estimated 3D geometry.
% \vspace{-1em}
To enable long-term memory in video generation, existing methods primarily fall into two categories. The first category, explicit 3D geometry-based methods~\cite{zhao2026spatia, yang2026neoverse, huang2025voyager, kong2025worldwarp, ren2025gen3c, wu2025spaMem}, incrementally reconstructs persistent 3D representations (e.g., point clouds, 3D Gaussians) during novel view generation. They employ reconstruction models~\cite{wang2025vggt, wang2025cut3r, wang2025moge} to estimate geometry and render partial target views, relying on generative models to inpaint missing regions. While conceptually sound, they suffer from inherent scale ambiguity during reconstruction, easily leading to misalignment between the estimated geometry and the user-defined camera trajectory in actual practice. This often causes inaccurate camera navigation and introduces revisit inconsistencies, as shown in Fig.~\ref{fig:intro} (top).

% \cyh{The second category include implicit multi-view-based methods~\cite{xiao2025worldmem, yu2025cam, sun2025worldplay}, which retrieves relevant historical frames based on camera FoV overlap to condition video generation.}

% The second category consists of implicit multi-view-based methods~\cite{xiao2025worldmem, yu2025cam, sun2025worldplay}, which retrieve relevant historical frames based on camera FoV overlap to condition video generation.
Instead of explicit reconstruction, the second category, implicit multi-view-based methods~\cite{xiao2025worldmem, yu2025cam, sun2025worldplay, wang2026matrix}, retrieves relevant historical frames based on camera Field-of-View (FoV) overlap to condition video generation. 
However, such FoV-based retrieval ignores geometric occlusions: frames whose visible regions are occluded from the target view may still be selected, as illustrated in Fig.~\ref{fig:intro} (bottom). This leads to visual inconsistencies and repeated content. While some works~\cite{li2025vmem, huang2025memoryForcing} attempt to mitigate this by reconstructing coarse 3D proxies for indexing, they reintroduce the scale estimation biases as in explicit methods.

% Some works~\cite{li2025vmem, huang2025memoryForcing} attempt to alleviate this issue by reconstructing coarse 3D proxies for indexing, but doing so reintroduces the scale estimation biases as in explicit geometry-based methods.

% The second category includes implicit multi-view-based methods~\cite{xiao2025worldmem, yu2025cam, sun2025worldplay}, which retrieve relevant historical frames based on camera FoV overlap to condition video generation.
% However, this naive FoV-based retrieval neglects geometric occlusions, as frames with occluded scenes can still be selected, as illustrated in Fig.~\ref{fig:intro} (top-right). This leads to visual inconsistencies and repeated content. While some works~\cite{li2025vmem, huang2025memoryForcing} attempt to mitigate this by reconstructing coarse 3D proxies for indexing, they inevitably reintroduce the scale estimation biases found in explicit methods.

In this work, we propose an implicit method for 3D-aware memory retrieval while avoiding the overhead and scale biases of explicit 3D reconstruction.
A key finding in our work is that the 3D priors of existing novel-view synthesis models can be utilized to help retrieve historical frames with significantly improved accuracy and occlusion awareness. 
In particular, the intermediate features of Feed-Forward Novel View Synthesis (FF-NVS) models~\cite{jin2024lvsm, jia2026efficient-lvsm} inherently encode rich 3D correspondence cues.
Building upon this, we propose a learning-based 3D-aware memory retrieval module. Specifically, we employ a lightweight Convolutional Neural Network (CNN) to process these features from the FF-NVS model and assess the view relevance of candidate frames to the target view. 
Instead of rule-based selection~\cite{yu2025cam, xiao2025worldmem}, our learning-based strategy captures the underlying 3D spatial relationships among generated images, allowing robust occlusion-aware retrieval without explicitly maintaining 3D geometry.

Retrieving correct frames is only half the challenge; effectively utilizing them is the other. Previous implicit methods~\cite{yu2025cam, xiao2025worldmem, sun2025worldplay} typically condition video generation directly on raw historical frames. This forces the network to learn complex 3D correspondences from scratch via computationally expensive full-parameter training. Furthermore, the absence of geometrically aligned guidance frequently leads to hallucinated repetitive visual content and degraded camera control. To address this, we propose an adaptive 3D-aligned memory injection mechanism. We employ a pre-trained FF-NVS module to warp the historical content to the target view, producing geometrically aligned guidance for the diffusion model. However, since NVS typically degrades under extrapolation and occlusion, we jointly fine-tune this module with the video diffusion model. This allows the system to emphasize reliable regions (mainly from interpolation) while suppressing unreliable extrapolated content. Consequently, the video diffusion model effectively benefits from the geometrically aligned guidance, achieving efficient training, superior generation consistency, and accurate camera control. 

In summary, our contributions are as follows: 
\begin{itemize} 
\item We propose a learning-based memory retrieval strategy that leverages implicit 3D-aware features to assess view relevance, enabling robust retrieval of relevant historical frames under occlusion without explicit 3D modeling.
\item We introduce an adaptive 3D-aligned memory injection module that implicitly aligns the retrieved frames to the target view while adaptively attending to reliable conditioning regions, thereby enhancing both revisit consistency and camera control precision for video generation.
\item Extensive experiments demonstrate that our method outperforms the state-of-the-art memory-conditioned video generation methods in terms of revisit consistency, generation fidelity, and camera control accuracy. 
\end{itemize}

% \item We introduce an aligned reliability-aware memory injection mechanism that provides explicit guidance for video generation. It adaptively attends to reliable condition regions while rectifying warping errors, thereby enhancing both revisit consistency and camera control precision.
% \item Extensive experiments validate that our method establishes a new state-of-the-art in memory-conditioned long-video generation, surpassing existing methods in terms of revisit consistency, generation fidelity, and camera control accuracy. 

% To summarize, our main contributions include:
% \begin{itemize}
%     \item We propose a learning-based 3D-aware memory retrieval module that utilizes implicit 3D features to score past views, enabling robust retrieval of 3D-relevant past views without explicit 3D geometry;
%     \item We design an aligned reliability-aware memory injection mechanism that aligns the retrieved past views to the target views while adaptively focusing on the reliable regions, achieving a good balance between camera control precision and video generation consistency;
%     \item Extensive experiments demonstrate that our method surpasses state-of-the-art open-source memory-conditioned long-video generation methods in revisit consistency, generation quality, and camera control accuracy.
% \end{itemize}

\section{Related Work}
\paragraph{Generalizable Novel View Synthesis.} Novel view synthesis (NVS) synthesizes unseen views of an underlying 3D scene from given source images. Existing generalizable NVS methods can be categorized into interpolation-based~\cite{chen2024mvsplat, jin2024lvsm, sajjadi2022srt, jiang2025anysplat, szymanowicz2026lagernvs} and extrapolation-based~\cite{yu2024viewcrafter, zhou2025seva, wang2024motionctrl, sargent2024zeronvs} approaches. Interpolation methods are typically deterministic, employing neural networks to predict explicit 3D representations~\cite{chen2024mvsplat,jiang2025anysplat} or directly regress target views~\cite{jin2024lvsm,sajjadi2022srt} in a feed-forward manner. While producing high-fidelity interpolated views, they inherently struggle to generate entirely new scene content. Instead, extrapolation methods rely on generative models, particularly camera-controlled video diffusion models~\cite{wang2024motionctrl,yu2024viewcrafter,bai2025recammaster,huang2025voyager}, to synthesize views beyond observed regions (also termed Video Scene Generation). However, they struggle to maintain global consistency with previously generated scenes over long durations.

% Conversely, extrapolation methods utilize generative models~\cite{sargent2024zeronvs,yu2024viewcrafter,zhou2025seva,wang2024motionctrl} to extend beyond original observations. To ensure temporal coherence, most works employ video diffusion models equipped with camera control~\cite{wang2024motionctrl,yu2024viewcrafter,bai2025recammaster,huang2025voyager} to extrapolate new content (also termed Video Scene Generation), yielding realistic novel views. However, they struggle to maintain global consistency with previously generated scenes over long durations.

\paragraph{Consistent Video Scene Generation.} To equip video scene generation with long-term consistency, memory mechanisms have been introduced, categorized into explicit 3D geometry-based and implicit multi-view-based approaches.

Explicit methods maintain a persistent 3D geometry for view re-projection and inpainting. Some works~\cite{ren2025gen3c, wu2025spaMem, zhao2026spatia, huang2025voyager} leverage off-the-shelf estimators~\cite{wang2025moge, wang2025cut3r, chen2025ttt3r, wang2025vggt} to construct point clouds, while others~\cite{yu2024viewcrafter, kong2025worldwarp, yang2026neoverse} further optimize 3D Gaussians to enhance visual quality. Nevertheless, these methods are bottlenecked by reconstruction errors and scale ambiguity, causing inaccurate navigation and severe inconsistencies upon scene revisits.

Implicit methods maintain a memory bank of previously generated frames, retrieving historical context based on camera FoV overlap~\cite{xiao2025worldmem, yu2025cam, sun2025worldplay}. However, this strategy ignores geometric occlusions, often selecting frames invisible from the target view. Some works~\cite{li2025vmem, huang2025memoryForcing} mitigate this by reconstructing coarse geometry, but they reintroduce scale biases. Retrieved frames are then used to condition video generation through cross-attention~\cite{xiao2025worldmem}, latent concatenation~\cite{yu2025cam,sun2025worldplay,wu2026infinite,wang2026matrix}, or a dedicated generative NVS model~\cite{li2025vmem}. However, such geometrically unaligned context tends to compromise camera control and produce repetitive, inconsistent results.

% To condition video generation on retrieved past frames, Worldmem~\cite{xiao2025worldmem} uses cross attention; CaM~\cite{yu2025cam} and WorldPlay~\cite{sun2025worldplay} concatenate historical context with target latents; Vmem~\cite{li2025vmem} employs a dedicated generative NVS model~\cite{zhou2025seva}. However, directly conditioning on such unaligned historical context tends to compromise camera control and produces repetitive, inconsistent results. 

To bridge this gap, our method introduces an implicit, 3D-aware memory mechanism, ensuring occlusion-robust consistency and accurate camera navigation without incurring the overhead of explicit 3D maintenance.

\section{Method}
\label{sec:method}

% \begin{wrapfigure}{r}{0.5\textwidth}
%   \centering
%   \includegraphics[width=0.48\textwidth]{fig/retrieval_v1.pdf}
%   \caption{3D-aware Memory Retrieval Module.}
%   \label{fig:memory_retrieval}
% \end{wrapfigure}

% \cyh{In this section, we present our framework for consistent video scene generation. First of all, we outline the preliminaries in Subsec.~\ref{sec:pre}. Afterwards, we elaborate on the details of our method by sequentially introducing an implicit \textit{3D-aware memory retrieval} module (Subsec.~\ref{sec:retriveal}) and a frames-dependent \textit{adaptive memory injection} mechanism (Subsec.~\ref{sec:injection}), as illustrated in Fig.~\ref{fig:method}.}

\begin{figure*}[t]
  \centering
  \includegraphics[width=0.9\textwidth]{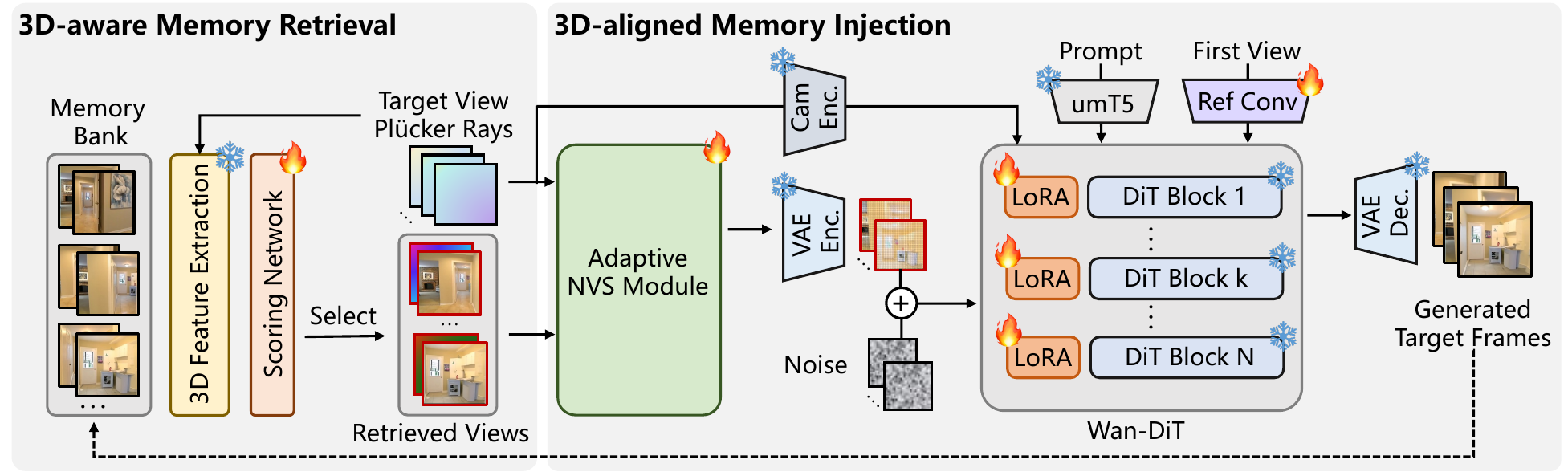}
  \caption{Overview of the proposed \textbf{I3DM} framework. Left: \textit{3D-aware Memory Retrieval.} For each historical frame in the memory bank, we first extract 3D-aware intermediate features using a pre-trained NVS model. A lightweight scoring network then evaluates their spatial relevance to the target view to select the most relevant frames. Right: \textit{3D-aligned Memory Injection.} The retrieved frames are processed by an Adaptive NVS Module to align them with the target view. These aligned results are then used to condition the Wan-DiT backbone for consistent video scene generation.}
  \label{fig:method}
% \vspace{-0.5em}
\end{figure*}

We introduce \textbf{I3DM} for consistent video scene generation, as shown in Fig.~\ref{fig:method}. I3DM comprises two key components: an implicit \textit{3D-aware memory retrieval} module (Sec.~\ref{sec:retriveal}) to select the most relevant historical frames, and an adaptive \textit{3D-aligned memory injection} module (Sec.~\ref{sec:injection}) to align the retrieved frames with the target view and condition the video diffusion model.

% In this section, we present our framework for consistent video scene generation, which comprises two key components as illustrated in Fig.~\ref{fig:method}.
% We first outline the preliminaries (Sec.~\ref{sec:pre}) and then introduce our implicit 3D-aware memory retrieval module (Sec.~\ref{sec:retriveal}). Subsequently, we detail our adaptive memory injection mechanism (Sec.~\ref{sec:injection}), which conditions the diffusion model on the retrieved frames.

% \mycfigure{method}{method_v3}{Overview of the proposed \textbf{I3DM} framework. Left: \textit{3D-aware Memory Retrieval.} We first extract 3D-aware intermediate features from a pre-trained NVS model for each historical frame in the memory bank. A lightweight scoring network then evaluates their spatial relevance to the target view to select the most relevant frames. Right: \textit{3D-aligned Memory Injection.} The retrieved frames are processed by an Adaptive NVS Module to align them with the target view. These aligned results are then used to condition the Wan-DiT backbone for consistent video scene generation.}

\subsection{Preliminaries}
\label{sec:pre}
\paragraph{Camera-Conditioned Video Generation.} 
Our framework builds upon Wan 2.1~\cite{wan2025wan}, a full-sequence latent video diffusion model comprising a causal 3D VAE and a Diffusion Transformer (DiT)~\cite{peebles2023scalable} denoiser. To enable camera control, we adopt a pre-trained camera-conditioned adaptation~\cite{he2024cameractrl} of the Wan model. 
Specifically, camera extrinsic and intrinsic parameters are encoded as 6D Plücker ray embeddings~\cite{plucker1865xvii} $\mathbf{P}$, which are mapped to camera features by a camera adapter and injected into video latents via element-wise addition for camera-controlled video generation.
\paragraph{Feed-forward Novel View Synthesis.} 
We adopt LVSM, a feed-forward NVS framework~\cite{jin2024lvsm}, to facilitate 3D-aware memory retrieval and injection. 
Given $N$ source images and their Plücker ray embeddings $\{(\mathbf{I}_i, \mathbf{P}_i)\}_{i=1}^N$, LVSM synthesizes a novel target view $\mathbf{I}^\text{t}$ by querying the target Plücker ray embedding $\mathbf{P}^\text{t}$. Each source RGB-ray pair is concatenated channel-wise, patchified, and linearly projected into source tokens $\{\mathbf{S}_i\}_{i=1}^N$, while the target rays are projected to target tokens $\mathbf{S}^\text{t}$ separately:
\begin{align}
    \{\mathbf{S}_i\}_{i=1}^N &= \mathrm{Linear}_\text{in}(\mathrm{Patchify}(\{(\mathbf{I}_i, \mathbf{P}_i)\}_{i=1}^N)), \label{eq:tokenize-source} \\
    \mathbf{S}^\text{t} &= \mathrm{Linear}_\text{tgt}(\mathrm{Patchify}(\mathbf{P}^\text{t})). \label{eq:tokenize-target}
\end{align}
The source and target tokens are then jointly processed by $L$ Transformer layers $\Phi$ to obtain target-view-aligned features:
\begin{equation}
    \mathbf{R}^\text{t} = \mathrm{\Phi}_{1 \to L}(\{\mathbf{S}_i\}_{i=1}^N, \mathbf{S}^\text{t}).
\end{equation}
Finally, the output features $\mathbf{R}^\text{t}$ are regressed to RGB values and unpatchfied to the synthesized novel view $\hat{\mathbf{I}}^\text{t}$:
\begin{equation}
\label{eq:nvs}
    \hat{\mathbf{I}}^\text{t} = \mathrm{Unpatchify}(\mathrm{Sigmoid}(\mathrm{Linear}_\text{out}(\mathbf{R}^\text{t}))).      
\end{equation}

\subsection{Implicit 3D-aware Memory Retrieval}
\label{sec:retriveal}

% To maintain scene consistency in long-term video generation, our first goal is to retrieve relevant historical frames that maximize scene overlap with the target view. 
% \cyh{However, relying on explicit 3D modeling for this task often introduces estimation biases and a simple FOV-based scheme struggles with occlusion-induced missing regions.  }
% To achieve 3D-aware retrieval (robust to complex occlusions) without incurring the estimation bias of explicit 3D modeling, our idea is to leverage the intermediate features of a pre-trained feed-forward NVS model~\cite{jin2024lvsm} that inherently encodes rich 3D correspondences. Specifically, we propose an implicit 3D-aware \cyh{memory} retrieval module (illustrated in Fig.~\ref{fig:memory_retrieval} \cyh{left}) that utilizes these features to select optimal historical frames by naturally reasoning about spatial occlusions.

\begin{figure}[t]
  \centering
  \includegraphics[width=1.0\columnwidth]{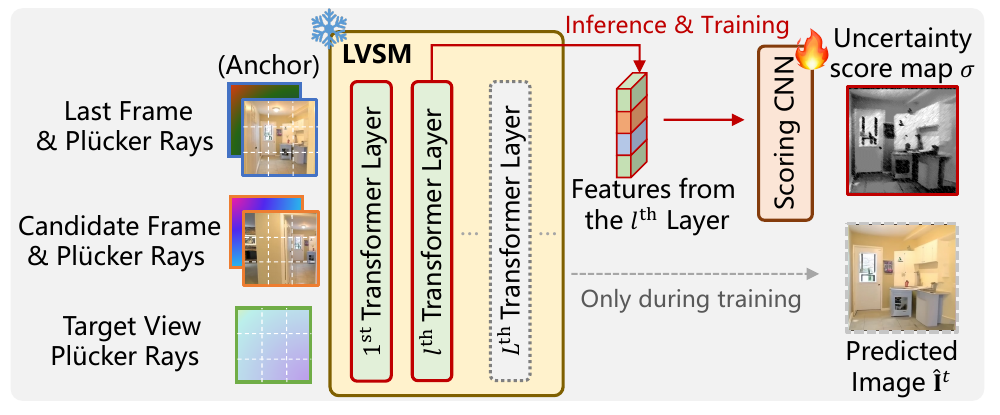}
  \caption{3D-aware Memory Retrieval Module. We first set the last frame as an anchor. For each historical candidate frame, we extract implicit 3D-aware features from the $l^{\text{th}}$ layer of the frozen LVSM. Based on these features, a Scoring CNN then predicts a spatial uncertainty map for the target view. Notably, the full Transformer is executed \textit{only during training} to provide supervision; during inference, the process terminates at the $l^{\text{th}}$ layer to efficiently deduce the score map.}
  \label{fig:memory_retrieval}
% \vspace{-0.5em}
\end{figure}

To maintain scene consistency in long-term video generation, our first goal is to retrieve relevant historical frames that maximize scene overlap with the target view. However, relying on explicit 3D modeling for this task often introduces estimation biases, and a simple FoV-based scheme fails under complex occlusions. To overcome these challenges, our idea is to leverage the intermediate features of a pre-trained feed-forward NVS model~\cite{jin2024lvsm}, which inherently encode rich 3D correspondences. Specifically, we propose an implicit \textit{3D-aware memory retrieval} module (illustrated in Fig.~\ref{fig:memory_retrieval}) that utilizes these features to select optimal historical frames by naturally reasoning about spatial occlusions.

% $\mathcal{V}=\{(\mathbf{I}_i, \mathbf{P}_i)\}_{i=1}^N$ 
\paragraph{3D-aware Scoring Network.} Let $\mathcal{V}=\{(\mathbf{I}_i, \mathbf{P}_i)\}_{i=1}$ denote the memory bank containing \textit{historical} frames and their Plücker ray embeddings. Given a target view with Plücker rays $\mathbf{P}^\text{t}$, our goal is to select a subset of geometrically relevant frames from $\mathcal{V}$. As the last frame typically preserves significant visual overlap with the target view, we mandatorily set it as an anchor. For each remaining frame in the memory bank, we treat it as a candidate and evaluate its relevance to the target view by comparing it with the anchor frame.

Specifically, we tokenize the last and candidate frames into $\mathbf{S}^{\mathrm{last}}$ and $\mathbf{S}^{\mathrm{cand}}$ using Eq.~\eqref{eq:tokenize-source}, and the target-view rays into $\mathbf{S}^{\mathrm{t}}$ using Eq.~\eqref{eq:tokenize-target}. We then feed these tokens into the frozen LVSM Transformer. Instead of executing full Transformer layers, we only extract intermediate features from the shallow $l^{\text{th}}$ layer. These shallow-layer features are capable of assessing multi-view relevance while maintaining lightweight computation during inference. Then, a lightweight, trainable scoring CNN decodes them to a spatial uncertainty score map:
\begin{equation}
    \mathbf{\sigma} = \text{CNN}_{\theta}(\mathrm{\Phi}_{1 \to l}(\mathbf{S}^{\mathrm{last}}, \mathbf{S}^{\mathrm{cand}}, \mathbf{S}^\text{t})),
\end{equation}
where $\mathbf{\sigma}\in\mathbb{R}^{\frac{H}{p}\times\frac{W}{p}}$ represents the spatial uncertainty of synthesizing the target frame using this specific candidate frame and the last frame. Here, $H \times W$ denotes the image resolution of the candidate frame and $p$ is the patch size.

\paragraph{Training.} To train the scoring CNN without explicit 3D ground truth, we draw inspiration from prior visual geometry work~\cite{wang2024dust3r,kendall2017uncertainties} and optimize an uncertainty loss $\mathcal{L}_{\text{un}}$ for each candidate frame:
\begin{multline}
    \mathcal{L}_{\text{un}} = \frac{1}{2}\sum_{u,v}
    \biggl(e^{-\mathbf{\sigma}(u,v)}\cdot \\
    \text{sg}\!\left[\mathrm{MSE}\!\left(
    \mathbf{\hat{I}}^\text{t}(u,v),\mathbf{I}^\text{t}(u,v)\right)\right]
    + \mathbf{\sigma}(u,v) \biggr),
\end{multline}
where $\mathbf{\sigma}(u,v)$, $\mathbf{\hat{I}}^\text{t}(u,v)$, and $\mathbf{I}^\text{t}(u,v)$ are the predicted uncertainty map, synthesized NVS image, and ground truth image for patch $(u,v)$. $\text{sg}[\cdot]$ denotes the stop-gradient operation. The intuition is that the pre-trained NVS model naturally synthesizes high-fidelity image patches in observed regions (yielding low MSE and driving $\mathbf{\sigma}(u,v)$ down), while producing blurry results in unobserved or occluded regions. Minimizing this loss encourages the network to produce high uncertainty $\mathbf{\sigma}(u,v)$ in unreliable regions. Note that we only execute the full LVSM Transformer during this training phase to predict $\mathbf{\hat{I}}^\text{t}$ for loss calculation.

% \begin{equation}
%     \mathcal{L}_{\text{un}} = \frac{1}{M} \sum_{j=1}^{M} \left( \frac{1}{2}e^{-\mathbf{u}_j}\cdot \text{sg}[{\mathrm{MSE}(\mathbf{\hat{I}}^\text{t}_j,\mathbf{I}^\text{t}_j)}] + \frac{1}{2} \mathbf{u}_j \right),
% \end{equation}
% where $\mathbf{u}_j$, $\mathbf{\hat{I}}^\text{t}_j$, and $\mathbf{I}^\text{t}_j$ are the predicted uncertainty map, synthesized NVS image, and ground truth image for the $j^{\text{th}}$ image patch, respectively. \cyh{delete: $M$ is the total number of patches, and} $M$ is the total number of patches, and
 % To retrieve $K$ total frames (excluding the mandatory last view),
 
% \paragraph{Inference.} During inference, we define the spatial confidence map for each historical candidate frame as its negative predicted uncertainty, $\mathbf{m} := -\mathbf{\sigma}$. To retrieve $K$ additional reference frames besides the last frame ($K+1$ in total), a naive approach would be a Top-$K$ selection based on the spatially averaged $\mathbf{m}$. However, this often yields redundant information, as temporally adjacent frames always share similar high-confidence regions.

\paragraph{Inference.} During inference, we define the spatial confidence map for each historical candidate frame as its negative predicted uncertainty, $\mathbf{m} := -\mathbf{\sigma}$. To retrieve $K$ additional reference frames ($K+1$ including the last frame), a naive way would be Top-$K$ selection based on the spatially averaged $\mathbf{m}$. However, this often yields severe redundancy, as temporally adjacent frames share similar high-confidence regions.

To promote broader scene coverage for a target view, we formulate the reference frame selection as a \textit{greedy maximum coverage problem}.
We maintain a global confidence canvas $\mathbf{m}^{\mathrm{g}}\in\mathbb{R}^{\frac{H}{p}\times\frac{W}{p}}$, initialized to zero, which aggregates the coverage from selected reference frames. Our goal is to select $K$ reference frames whose confidence maps collectively maximize the $\mathbf{m}^{\mathrm{g}}$ via a patch-wise maximization operation. 

% Let $\mathcal{C}$ denote the set of currently selected frames, initialized to $\emptyset$. At each iteration, we select the candidate frame that yields the largest information gain for the global confidence canvas and add it to $\mathcal{C}$. 
% Specifically, for a candidate frame $i$ with confidence map $\mathbf{m}_i$ in the memory bank $\mathcal{V}$, we compute the updated canvas via a patch-wise maximum operation, and the information gain of frame $i$ is defined as the total improvement over the current canvas $\mathbf{m}^{\mathrm{g}}$:
% \begin{equation}
%     i^* = \underset{i \notin \mathcal{C}}{\arg\max} 
%     \sum_{u,v} 
%     \Big(
%     \max\left(\mathbf{m}^{\mathrm{g}}(u,v) , \mathbf{m}_i(u,v) \right)
%     - 
%     \mathbf{m}^{\mathrm{g}}(u,v)
%     \Big).
% \end{equation}

Let $\mathcal{C}$ denote the set of currently selected frame indices, initialized to $\emptyset$. At each iteration, for each unselected candidate frame $i$ in the memory bank $\mathcal{V}$, excluding the anchor frame, we evaluate the marginal gain over the current global confidence $\mathbf{m}^{\mathrm{g}}$ through a patch-wise maximum operation. We then select the candidate index $i^*$ with the largest gain:
\begin{equation}
    i^* =
    \underset{i \notin \mathcal{C}}{\arg\max}
    \sum_{u,v}
    \left[
    \max\left(\mathbf{m}^{\mathrm{g}}(u,v),\mathbf{m}_i(u,v)\right)
    - \mathbf{m}^{\mathrm{g}}(u,v)
    \right].
\end{equation}
We then update the selected frame set $\mathcal{C}\leftarrow\mathcal{C}\cup\{i^*\}$ and global confidence $\mathbf{m}^{\mathrm{g}}(u,v) \leftarrow \max\!\left(\mathbf{m}^{\mathrm{g}}(u,v),\, \mathbf{m}_{i^*}(u,v)\right)$. This procedure repeats until $K$ reference frames are selected. By evaluating incremental patch-wise gains, this redundancy-aware strategy promotes complementary coverage among selected candidates. Together, the last frame preserves local temporal continuity, while the selected historical frames provide additional context for target regions insufficiently observed in the last frame.

When processing a sequence of $T$ target views, we maintain one global confidence canvas for each view. At each iteration, we select the candidate frame that maximizes the average marginal gain across ALL $T$ canvases. Since this iterative greedy maximum coverage search only uses simple operations (e.g., patch-wise maximum and average) executed at a reduced patch resolution, it incurs negligible inference overhead. Algorithm details are provided in the Appendix.

\subsection{Adaptive 3D-aligned Memory Injection}
\label{sec:injection}

Having retrieved geometrically relevant historical frames in $\mathcal{C}$, the next challenge is to effectively condition the video diffusion model. Directly injecting unaligned historical frames forces the diffusion backbone to learn complex spatial transformations from scratch, often causing degraded generation consistency and repetitive visual content. To address this, we introduce a 3D-aligned memory injection mechanism (Fig.~\ref{fig:method}, right) that leverages a pre-trained NVS module to adaptively produce spatially aligned guidance for the diffusion model.

% \cyh{\paragraph{3D-aligned guidance.}} ?
\paragraph{Adaptive 3D Alignment.} Given the retrieved frame set $\mathcal{C}$ and the last frame, we use the pre-trained LVSM~\cite{jin2024lvsm} to warp them to the target view following Eq.~\ref{eq:nvs}. The key idea is to leverage the rich 3D priors inherently encoded in the pre-trained LVSM as a powerful geometric scaffold for accurate spatial alignment. This design decouples complex 3D transformations from the generative process, allowing the diffusion model to focus entirely on its core strength: generating new content in unobserved regions.

% The key advantage of this design is that the pre-trained LVSM inherently encodes rich 3D priors, serving as a powerful geometric scaffold for rigorous spatial alignment. This module decouples complex 3D transformations from the generative process, allowing the diffusion model to focus entirely on its core strength: generating new content in unobserved regions.

% The key idea is to leverage the rich 3D priors inherently encoded in the pre-trained LVSM as a powerful geometric scaffold for accurate spatial alignment. 

% However, the pre-trained LVSM produces reliable results only in observed interpolated regions and introduces warping artifacts in unobserved extrapolated regions, interfering with subsequent generation. 

However, the pre-trained LVSM is reliable only when interpolating within observed regions and introduces warping artifacts when extrapolating into unobserved regions, interfering with subsequent generation. Although explicit 3D memory methods also provide 3D-aligned guidance, they rely on rigid, non-differentiable representations (e.g., point clouds). In contrast, our fully neural architecture enables \textit{joint fine-tuning} of the NVS module with the diffusion model. This crucial design shifts the NVS objective from strict photometric reconstruction to learning optimal conditioning features. The fine-tuned NVS module learns to produce sharp, accurate content observed in $\mathcal{C}$ while providing soft, uncertainty-aware features in extrapolated areas. 
This enables the downstream diffusion model to adaptively attend to the guidance: relying on high-fidelity aligned regions for strict consistency, while downweighting uncertain areas to fall back on internal generative priors (see Appendix for examples).

\paragraph{Memory-Conditioned Video Generation.}
We encode the adaptively aligned memory $\hat{\mathbf{I}}^\text{t}$ using the VAE encoder, yielding the latent condition $\mathbf{z}^{\text{mem}}$. It is channel-wise concatenated with the noisy latent $\mathbf{z}_t$ as $\mathbf{z}^\prime=[\mathbf{z}_t,\mathbf{z}^{\text{mem}}]$ and fed into the LoRA-equipped DiT blocks for conditioned generation (Fig.~\ref{fig:method}). To preserve pre-trained generative priors, we freeze the VAE, text encoder, camera adapter, and original DiT weights. For global temporal coherence, the last frame is injected as the first frame of each clip through a trainable reference convolutional layer. We train the framework with the flow-matching loss~\cite{lipman2022flow}. By back-propagating the generation loss through the VAE to the NVS module, the system adaptively learns optimal conditioning signals for the best generative outcome.

\section{Experiments}

% In this section, we first detail the experimental setup of our proposed I3DM framework. We then compare our approach with state-of-the-art camera-controlled video generation methods. Furthermore, we perform ablation studies to validate the effectiveness of our memory retrieval strategy, the proposed memory injection mechanism, and the choice of different feature layers for retrieval.

In this section, we first detail the experimental setup of \textbf{I3DM} (Sec.~\ref{sec:setup}). We then compare our approach with state-of-the-art camera-controlled video generation methods (Sec.~\ref{sec:comp}). Furthermore, we perform ablation studies to validate the effectiveness of our memory retrieval and injection strategies (Sec.~\ref{sec:ab}). More experiments are provided in our Appendix.

\subsection{Experimental Setup}
\label{sec:setup}

% \paragraph{Datasets.} We evaluate on two public real-world datasets: \sloppy RealEstate10K (Re10K)~\cite{zhou2018stereo} and Tanks-and-Temples (T\&T)~\cite{Knapitsch2017tnt}, both containing diverse indoor and outdoor scenes with camera annotations. For text conditioning, we generate short captions for all video clips using Qwen2.5-VL-7B~\cite{qwen2.5-VL}. We train our model solely on the Re10K training set and evaluate on the Re10K test set and T\&T dataset to assess out-of-distribution generalization.

\paragraph{Datasets.} We evaluate on two public real-world datasets: \sloppy RealEstate10K (Re10K)~\cite{zhou2018stereo} and Tanks-and-Temples (T\&T)~\cite{Knapitsch2017tnt}, both containing diverse indoor and outdoor scenes with camera annotations. We caption all video clips for text conditioning using Qwen2.5-VL-7B~\cite{qwen2.5-VL}. We only train our model on the Re10K training set, while evaluating on the Re10K test set and T\&T dataset to assess out-of-distribution generalization.

\paragraph{Implementation Details.} 
(1) Training of Memory Retrieval. We use the pre-trained LVSM~\cite{jin2024lvsm} as the feature extractor. For efficiency, input images are resized to $256\times256$, and features are extracted from the $6^{\text{th}}$ Transformer layer. These features are processed by a 3-layer scoring CNN with channels 256, 64, and 1. We train this module using a learning rate of $5 \times 10^{-5}$ and batch size of 64. 
(2) Training of Video Generation. We use Wan-CamCtrl-1.3B~\cite{wan2025wan} as the backbone and train on 77-frame clips at $640\times352$ resolution, conditioned on 4 reference frames randomly sampled from the full video sequence. The Adaptive NVS Module is initialized from the pre-trained LVSM. We jointly fine-tune the NVS Transformer blocks, the reference convolution layer, and the DiT LoRA layers for 11k steps with batch size 4. 
(3) Long Video Inference. We generate long videos autoregressively, one clip at a time. For each 77-frame target clip, we uniformly sample 20 views as retrieval queries. Using our retrieval module, we select $K=3$ relevant historical frames and the last frame from the memory bank. After each clip, every \textit{fourth} generated frame is added to the memory bank to support subsequent generation.

\paragraph{Evaluation Metrics.}
We assess performance across three dimensions:
(1) Video Generation Quality. We use Fréchet Inception Distance (FID)~\cite{heusel2017fid} and Fréchet Video Distance (FVD)~\cite{unterthiner2018fvd} to measure distributional divergence from ground truth, and the Imaging Quality (IMQ) metric from VBench~\cite{huang2023vbench} when ground truth is unavailable. 
(2) Camera Control Precision. We estimate camera poses of generated videos using Pi3~\cite{wang2025pi}, and report rotation ($R_{err}^\circ$) and translation ($T_{err}$) errors against the ground-truth trajectory. Both trajectories are aligned by setting the first frame as reference and normalizing translation scale using the farthest frame.
(3) Revisit Consistency. Following Vmem~\cite{li2025vmem}, we use forward-and-reverse camera trajectories and compare generated frames at the same viewpoint using PSNR, SSIM~\cite{wang2004ssim}, and LPIPS~\cite{zhang2018lpips}.

\begin{table}[t]
    \centering
    \footnotesize
    \caption{
        Comparisons on Re10K (top) and T\&T (bottom).
    }
    \label{tab:comp_re10k_tnt}

    \setlength{\aboverulesep}{0pt}
    \setlength{\belowrulesep}{0pt}
    \setlength{\tabcolsep}{2.8pt}
    \renewcommand{\arraystretch}{1.03}

    \resizebox{\columnwidth}{!}{%
    \begin{tabular}{
        @{}l
        @{\hspace{4pt}}cc
        @{\hspace{6pt}}cc
        @{\hspace{6pt}}ccc
        @{}
    }
        \toprule

        \multirow{2}{*}{\textbf{Method}}
        & \multicolumn{2}{c}{\textbf{Visual Quality}}
        & \multicolumn{2}{c}{\textbf{Camera Control}}
        & \multicolumn{3}{c}{\textbf{Revisit Consistency}} \\

        \cmidrule(lr){2-3}
        \cmidrule(lr){4-5}
        \cmidrule(lr){6-8}

        & \textbf{FID}$\downarrow$
        & \textbf{FVD}$\downarrow$
        & $\boldsymbol{R_{\mathrm{err}}^\circ}\!\downarrow$
        & $\boldsymbol{T_{\mathrm{err}}}\!\downarrow$
        & \textbf{PSNR}$\uparrow$
        & \textbf{SSIM}$\uparrow$
        & \textbf{LPIPS}$\downarrow$ \\

        \midrule
        \multicolumn{8}{c}{\textbf{Re10K Dataset}} \\
        \midrule

        ViewCrafter
        & 58.63
        & --
        & 71.116
        & 0.7867
        & 9.29
        & 0.270
        & 0.6797 \\

        Voyager
        & 31.90
        & 377.86
        & 17.801
        & 0.3471
        & 14.75
        & 0.499
        & 0.4708 \\
        
        Gen3C
        & 29.05
        & 292.56
        & \cellcolor{thirdcolor}8.183
        & \cellcolor{secondcolor}0.1450
        & 15.23
        & 0.553
        & 0.4397 \\

        WorldWarp
        & \cellcolor{secondcolor}21.90
        & \cellcolor{secondcolor}181.95
        & 13.003
        & \cellcolor{thirdcolor}0.2161
        & 14.77
        & 0.592
        & 0.4828 \\

        WorldPlay
        & 24.14
        & 192.52
        & 15.298
        & 0.2848
        & \cellcolor{thirdcolor}15.71
        & 0.560
        & \cellcolor{thirdcolor}0.3878 \\

        NeoVerse
        & 30.69
        & 222.68
        & 21.140
        & 0.4114
        & 15.48
        & \cellcolor{thirdcolor}0.605
        & 0.4554 \\

        LB-World-F
        & 27.58
        & 327.04
        & 17.331
        & 0.4327
        & 13.46
        & 0.453
        & 0.4742 \\

        Inf-World
        & 24.86
        & 213.55
        & 21.776
        & 0.4610
        & 14.66
        & 0.476
        & 0.4136 \\

        Spatia
        & \cellcolor{thirdcolor}22.47
        & \cellcolor{thirdcolor}185.66
        & 15.979
        & 0.3077
        & 15.35
        & 0.534
        & 0.4680 \\

        VMem
        & 33.73
        & --
        & \cellcolor{secondcolor}6.782
        & 0.3615
        & \cellcolor{secondcolor}23.97
        & \cellcolor{secondcolor}0.727
        & \cellcolor{secondcolor}0.1670 \\

        Ours
        & \cellcolor{bestcolor}\textbf{15.56}
        & \cellcolor{bestcolor}\textbf{118.27}
        & \cellcolor{bestcolor}\textbf{2.044}
        & \cellcolor{bestcolor}\textbf{0.0507}
        & \cellcolor{bestcolor}\textbf{24.76}
        & \cellcolor{bestcolor}\textbf{0.828}
        & \cellcolor{bestcolor}\textbf{0.0743} \\

        \midrule
        \midrule

        & \textbf{FID}$\downarrow$
        & \textbf{IMQ}$\uparrow$
        & $\boldsymbol{R_{\mathrm{err}}^\circ}\!\downarrow$
        & $\boldsymbol{T_{\mathrm{err}}}\!\downarrow$
        & \textbf{PSNR}$\uparrow$
        & \textbf{SSIM}$\uparrow$
        & \textbf{LPIPS}$\downarrow$ \\

        \midrule
        \multicolumn{8}{c}{\textbf{T\&T Dataset}} \\
        \midrule

        Voyager
        & 166.76
        & 61.87
        & 33.751
        & 0.2980
        & 12.85
        & 0.278
        & 0.5644 \\
        
        Gen3C
        & 113.62
        & 57.55
        & \cellcolor{thirdcolor}6.853
        & \cellcolor{secondcolor}0.0862
        & \cellcolor{thirdcolor}17.96
        & \cellcolor{thirdcolor}0.519
        & \cellcolor{thirdcolor}0.2910 \\

        WorldWarp
        & 113.76
        & 64.52
        & 9.691
        & \cellcolor{thirdcolor}0.1828
        & 15.43
        & 0.442
        & 0.4577 \\

        WorldPlay
        & \cellcolor{bestcolor}\textbf{96.00}
        & \cellcolor{bestcolor}\textbf{73.94}
        & 24.751
        & 0.6214
        & 12.90
        & 0.334
        & 0.4575 \\

        NeoVerse
        & 168.05
        & 63.62
        & 29.154
        & 0.4953
        & 12.14
        & 0.301
        & 0.5884 \\

        LB-World-F
        & 155.14
        & 67.97
        & 33.811
        & 0.3784
        & 10.53
        & 0.262
        & 0.5984 \\

        Inf-World
        & 160.26
        & \cellcolor{thirdcolor}70.32
        & 36.001
        & 0.6682
        & 11.34
        & 0.250
        & 0.6027 \\

        Spatia
        & \cellcolor{thirdcolor}106.69
        & 66.01
        & 26.438
        & 0.6161
        & 13.56
        & 0.317
        & 0.4635 \\

        VMem
        & 128.25
        & --
        & \cellcolor{secondcolor}5.819
        & 0.6301
        & \cellcolor{bestcolor}\textbf{21.38}
        & \cellcolor{secondcolor}0.562
        & \cellcolor{secondcolor}0.1994 \\
        
        Ours
        & \cellcolor{secondcolor}96.26
        & \cellcolor{secondcolor}70.75
        & \cellcolor{bestcolor}\textbf{2.793}
        & \cellcolor{bestcolor}\textbf{0.0750}
        & \cellcolor{secondcolor}21.24
        & \cellcolor{bestcolor}\textbf{0.674}
        & \cellcolor{bestcolor}\textbf{0.0997} \\

        \bottomrule
    \end{tabular}%
    }

    % \vspace{-0.5em}
\end{table}

\begin{table}[t]
    \centering
    \footnotesize
    \caption{
        Ablation studies of memory retrievals (top) and memory injection mechanisms (bottom) on Re10K Dataset.
    }
    \label{tab:re10k_ablation}

    \setlength{\aboverulesep}{0pt}
    \setlength{\belowrulesep}{0pt}
    \setlength{\tabcolsep}{3.2pt}
    \renewcommand{\arraystretch}{1.08}

    \resizebox{\columnwidth}{!}{%
    \begin{tabular}{
        @{}l
        @{\hspace{5pt}}cc
        @{\hspace{7pt}}cc
        @{\hspace{7pt}}ccc
        @{}
    }
        \toprule

        \textbf{Method}
        & \textbf{FID}$\downarrow$
        & \textbf{FVD}$\downarrow$
        & $\boldsymbol{R_{\mathrm{err}}^\circ}\!\downarrow$
        & $\boldsymbol{T_{\mathrm{err}}}\!\downarrow$
        & \textbf{PSNR}$\uparrow$
        & \textbf{SSIM}$\uparrow$
        & \textbf{LPIPS}$\downarrow$ \\

        \midrule

        Temporal
        & 23.67 & 189.39 & 5.617 & 0.1277
        & 13.78 & 0.522 & 0.4660 \\
        
        Random
        & 18.50 & 141.88 & 2.075 & 0.0566
        & 19.25 & 0.687 & 0.2027 \\
        
        FoV-based
        & \cellcolor{thirdcolor}17.63
        & 134.69
        & \cellcolor{bestcolor}\textbf{1.895}
        & \cellcolor{thirdcolor}0.0510
        & \cellcolor{secondcolor}22.72
        & \cellcolor{secondcolor}0.781
        & \cellcolor{secondcolor}0.1084 \\
        
        Geo-based
        & 17.79
        & \cellcolor{secondcolor}133.66
        & \cellcolor{thirdcolor}2.027
        & 0.0511
        & 20.94
        & 0.738
        & 0.1346 \\
        
        I3D-TopK
        & \cellcolor{secondcolor}17.59
        & \cellcolor{thirdcolor}134.13
        & 2.046
        & \cellcolor{bestcolor}\textbf{0.0498}
        & \cellcolor{thirdcolor}22.33
        & \cellcolor{thirdcolor}0.769
        & \cellcolor{thirdcolor}0.1138 \\
        
        I3D-Ours
        & \cellcolor{bestcolor}\textbf{17.55}
        & \cellcolor{bestcolor}\textbf{131.66}
        & \cellcolor{secondcolor}1.991
        & \cellcolor{secondcolor}0.0505
        & \cellcolor{bestcolor}\textbf{24.73}
        & \cellcolor{bestcolor}\textbf{0.828}
        & \cellcolor{bestcolor}\textbf{0.0756} \\
        
        \midrule
        \midrule
        
        w/o mem.
        & \cellcolor{thirdcolor}21.43
        & \cellcolor{thirdcolor}169.28
        & \cellcolor{thirdcolor}6.096
        & \cellcolor{thirdcolor}0.1551
        & 12.73
        & \cellcolor{thirdcolor}0.516
        & 0.4910 \\
        
        w/o align.
        & 43.12
        & 314.40
        & 16.399
        & 0.2554
        & \cellcolor{thirdcolor}15.14
        & \cellcolor{secondcolor}0.568
        & \cellcolor{thirdcolor}0.4666 \\
        
        w/ fz. NVS
        & \cellcolor{bestcolor}\textbf{16.02}
        & \cellcolor{bestcolor}\textbf{121.56}
        & \cellcolor{secondcolor}1.993
        & \cellcolor{secondcolor}0.0591
        & \cellcolor{secondcolor}24.46
        & \cellcolor{bestcolor}\textbf{0.828}
        & \cellcolor{secondcolor}0.0760 \\
        
        w/ ft. NVS
        & \cellcolor{secondcolor}17.55
        & \cellcolor{secondcolor}131.66
        & \cellcolor{bestcolor}\textbf{1.991}
        & \cellcolor{bestcolor}\textbf{0.0505}
        & \cellcolor{bestcolor}\textbf{24.73}
        & \cellcolor{bestcolor}\textbf{0.828}
        & \cellcolor{bestcolor}\textbf{0.0756} \\

        \bottomrule
    \end{tabular}%
    }
\end{table}

\begin{figure*}[!t]
  \centering
  \includegraphics[width=\textwidth]{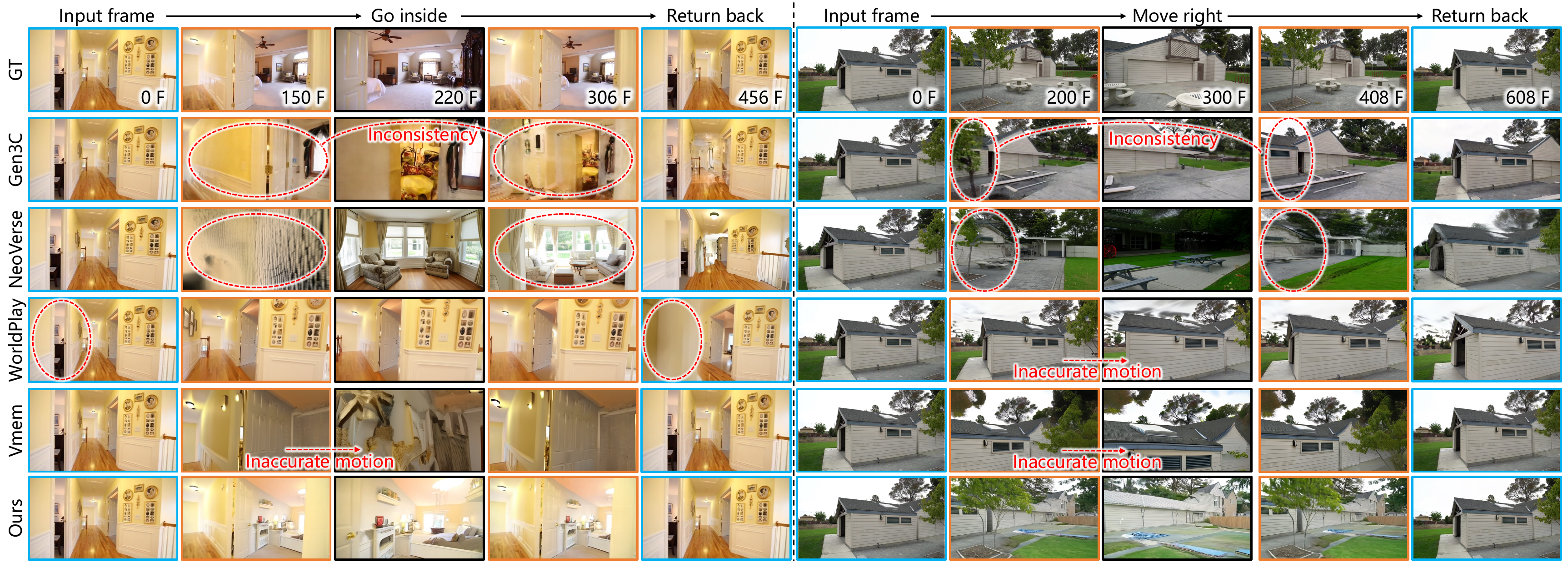}
  \caption{Qualitative comparison on the Re10K (left) and T\&T (right) datasets. The frames with matching colors should remain strictly consistent; red circles and arrows highlight visual inconsistencies and inaccurate camera motion, respectively.}
  \label{fig:comp_re10k_tnt}
% \vspace{-0.5em}
\end{figure*}

\subsection{Comparisons}
\label{sec:comp}

\paragraph{Baseline Models.} We compare our method against ten representative open-source baselines with memory mechanisms: (1) Explicit methods: ViewCrafter~\cite{yu2024viewcrafter}, Gen3C~\cite{ren2025gen3c}, WorldWarp~\cite{kong2025worldwarp}, Voyager~\cite{huang2025voyager}, NeoVerse~\cite{yang2026neoverse} and Spatia~\cite{zhao2026spatia}; (2) Implicit methods: WorldPlay~\cite{sun2025worldplay}, LingBot-World-Fast~\cite{lingbot-world} and Infinity-World~\cite{wu2026infinite}; and (3) Hybrid methods: Vmem~\cite{li2025vmem}. We follow their official implementations and evaluate on the same datasets for fair comparison.

% \paragraph{Results on RealEstate10K.} We conduct evaluations on a subset of 200 randomly selected scenes from the Re10K test set. To assess long-term scene consistency during revisits, we adopt the cycle-trajectory evaluation protocol from Vmem~\cite{li2025vmem}. This involves executing the original test trajectory followed by immediately retracing the path in reverse, resulting in sequences of 456 frames per scene. Video generation quality and camera control metrics are computed over the entire sequence, while revisit consistency metrics are calculated between spatially corresponding frames from the original and reversed trajectories. For Vmem~\cite{li2025vmem}, which generates discrete frames, we compute metrics at a stride of 10 frames to remain consistent with its original paper. As shown at the top of Tab.~\ref{tab:comp_re10k_tnt}, our method outperforms all baselines across all metrics. Qualitative comparisons are provided at the top of Fig.~\ref{fig:comp_re10k_tnt}. Gen3C and WorldWarp suffer from severe inconsistencies in revisited regions, while WorldPlay mitigates this but still cannot maintain strict consistency. Vmem yields unsatisfactory results due to inaccurate camera control (e.g., colliding with walls). In contrast, our method produces convincing results with accurate camera control and well-preserved visual consistency in revisited areas.

\paragraph{Results on RealEstate10K.} Since we focus on long-horizon evaluation, we filter out scenes with fewer than 230 frames from the Re10K test set and test on 507 scenes. Following the cycle-trajectory protocol of Vmem~\cite{li2025vmem}, we randomly sample 228 consecutive frames per scene as the forward trajectory and append its reverse, yielding a 456-frame sequence. Video quality and camera control metrics are computed over the full sequence, while revisit consistency is measured between spatially aligned frame pairs from the forward and reverse trajectories. Since Vmem generates discrete frames, we evaluate it at its native stride of 10, following its original setting. As shown in Tab.~\ref{tab:comp_re10k_tnt} (top), our method outperforms all baselines across all metrics. Qualitative comparisons are shown in Fig.~\ref{fig:comp_re10k_tnt} (left). Gen3C and NeoVerse show severe inconsistencies in revisited regions. WorldPlay mitigates this but still fails to preserve strict consistency, while Vmem yields unsatisfactory results due to inaccurate camera control (e.g., colliding with walls). In contrast, our method produces convincing results with accurate camera control and well-preserved visual consistency in revisited areas.

\paragraph{Results on Tanks-and-Temples.} To evaluate generalization, we test on the T\&T dataset. Since it consists of discrete image collections rather than continuous video, we apply $15\times$ temporal interpolation between adjacent frames to synthesize smooth camera motion. We evaluate all six scenes and apply the same cycle-trajectory protocol on the first 304 interpolated frames, yielding a sequence of 608 frames per scene. As shown at the Tab.~\ref{tab:comp_re10k_tnt} (bottom) and Fig.~\ref{fig:comp_re10k_tnt} (right), our method achieves better results in camera control and revisit consistency. Although Vmem achieves slightly higher PSNR for consistency, it suffers from severe translation errors that restrict novel content generation (i.e., the camera fails to move as instructed). In contrast, our method accurately follows target camera trajectories to explore more novel regions while maintaining strict revisit consistency.

\begin{figure}[t]
  \centering
  \includegraphics[width=1.0\columnwidth]{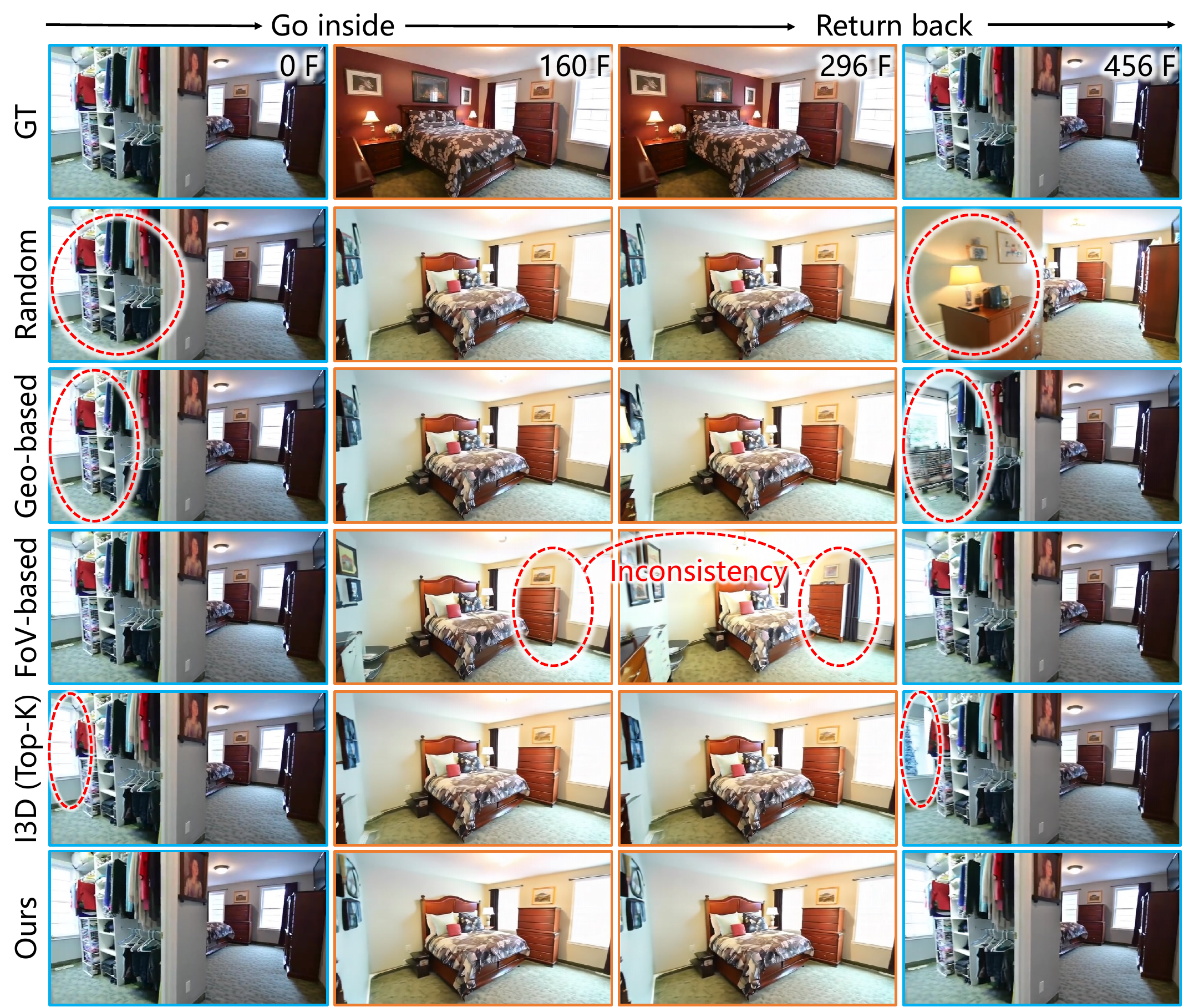}
  \caption{Ablation of memory retrieval strategies. Frames with the same color should remain strictly consistent. Red circles highlight visual inconsistencies. Temporal and random strategies fail to maintain scene consistency. Geometry-, FoV- and I3D-TopK-based approaches improve revisit consistency but still exhibit artifacts, while our full I3D-based retrieval strategy robustly maintains strict consistency.}
  \label{fig:ab_retrieval}
% \vspace{-0.5em}
\end{figure}

\begin{figure}[t]
  \centering
  \includegraphics[width=1.0\columnwidth]{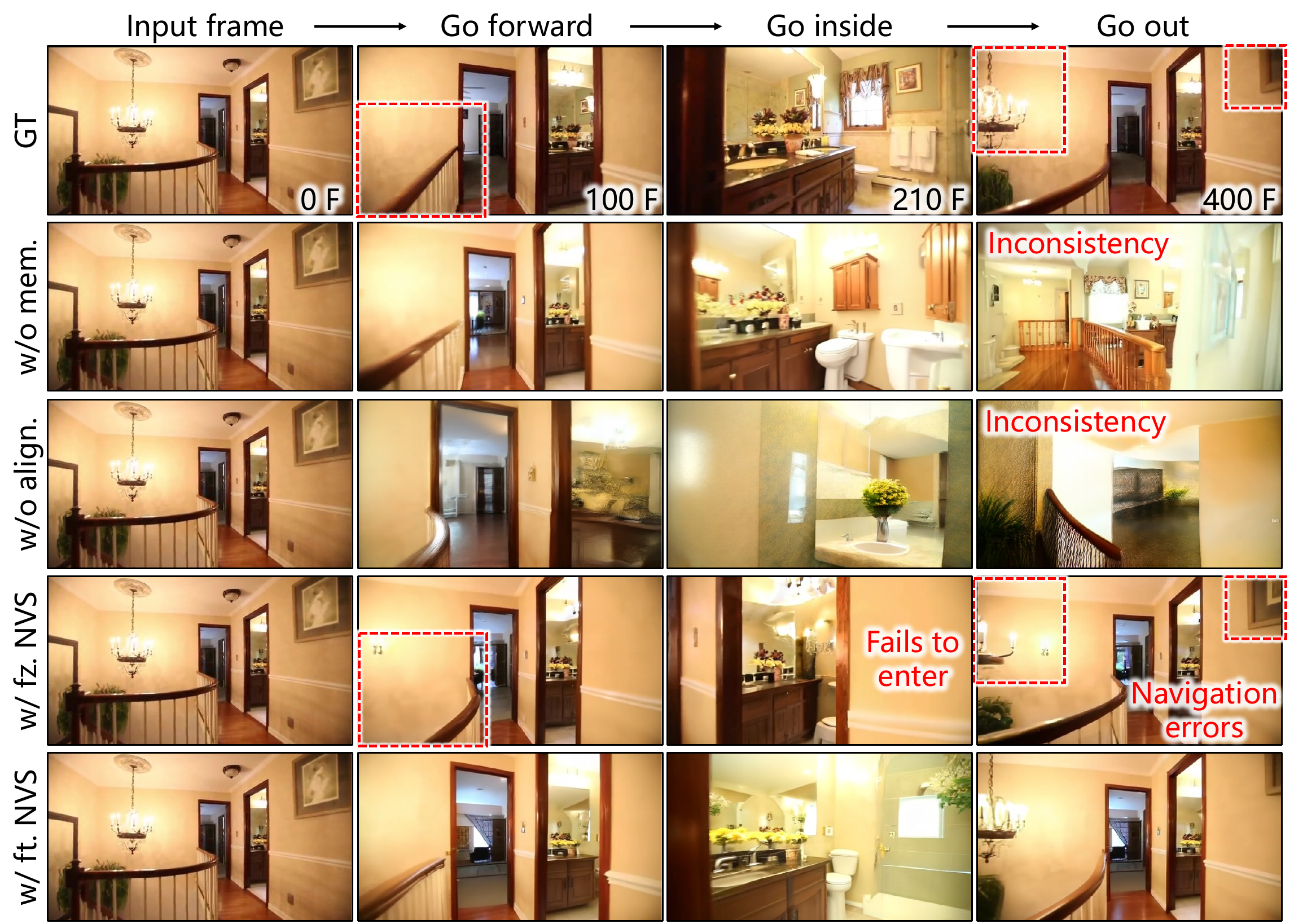}
  \caption{Ablation of memory injection mechanisms. Removing memory or spatial alignment compromises revisit consistency, while freezing the NVS module causes inaccurate camera motion and navigation failures (e.g., failing to enter the room). Jointly fine-tuning the NVS module ensures both consistent generation and accurate camera control.}
  \label{fig:ab_adaptive}
% \vspace{-0.5em}
\end{figure}

% \begin{table}[t]
%     \centering
%     \caption{Ablation studies of different feature layers for memory retrieval on the Re10K dataset. Time costs are averaged per video clip and reported in seconds (s).}
%     \label{tab:ab_layer}
    
%     \setlength{\aboverulesep}{0pt}
%     \setlength{\belowrulesep}{0pt}
    
%     \renewcommand{\arraystretch}{1.1}
    
%     \newcolumntype{S}{>{\hsize=0.9\hsize\centering\arraybackslash}X}
%     \newcolumntype{W}{>{\hsize=1.25\hsize\centering\arraybackslash}X}
    
%     \begin{tabularx}{\textwidth}{l | S S | S S S | W W}
%         \toprule
%         \multirow{2}{*}{Layer} & \multicolumn{2}{c|}{\textbf{Visual Quality}} & \multicolumn{3}{c|}{\textbf{Revisit Consistency}} & \multicolumn{2}{c}{\textbf{Per-clip Time Cost}} \\
        
%         \cmidrule(lr){2-3} \cmidrule(lr){4-6} \cmidrule(lr){7-8}
        
%         & FID $\downarrow$ & FVD $\downarrow$ & PSNR $\uparrow$ & SSIM $\uparrow$ & LPIPS $\downarrow$ & Retrieval $\downarrow$ & Generation $\downarrow$ \\
%         \midrule
        
%         L0  & 18.655 & 148.764 & 17.183 & 0.634 & 0.2764 & 0.1853 & 56.7241 \\
%         L4  & 18.098 & 133.384 & 24.609 & 0.823 & 0.0786 & 1.3481 & 58.3389 \\
%         L6  & 17.553 & 131.657 & 24.732 & 0.828 & 0.0756 & 1.9739 & 59.0871 \\
%         L12 & 17.656 & 134.054 & 24.771 & 0.828 & 0.0761 & 3.8737 & 61.4180 \\
%         L24 & 17.599 & 128.981 & 24.823 & 0.830 & 0.0742 & 7.6167 & 64.9117 \\
%         \bottomrule
%     \end{tabularx}
% \end{table}

\subsection{Ablation Study}
\label{sec:ab}

% We conduct ablation studies on the RealEstate10K dataset (Tab.~\ref{tab:re10k_ablation}) to validate our two key designs: the implicit 3D-aware memory retrieval strategy and the adaptive memory injection module. And we also 

\paragraph{Memory Retrieval Strategy.} We evaluate our implicit 3D-aware memory retrieval strategy against five alternatives: 
(1) temporal retrieval, selecting $K$ the most recent views; 
(2) random retrieval, selecting $K$ views randomly; 
(3) FoV-based retrieval, selecting the top-$K$ views by target-view FoV overlap, following Worldmem~\cite{xiao2025worldmem}; 
(4) geometry-based retrieval, selecting the $K$ most frequently referenced views using surfel-indexed retrieval and non-maximum suppression, as in Vmem~\cite{li2025vmem}; 
(5) I3D-based (TopK) retrieval, selecting the top-$K$ views with the highest averaged confidence score using implicit 3D priors. 
To ensure fair comparison, all methods share the same generation backbone and differ only in the retrieval. As reported in Tab.~\ref{tab:re10k_ablation} (top), our method outperforms alternatives on almost all metrics, improving revisit PSNR by 2.02 dB over the second-best baseline. Qualitatively (Fig.~\ref{fig:ab_retrieval}), temporal and random retrieval fail to maintain consistency during revisits. Geometry-based retrieval improves consistency but still shows artifacts, while FoV-based retrieval struggles under occlusions and fails to retrieve correct historical content. I3D-based Top-K selection still exhibits inconsistencies. In contrast, our full implicit 3D-aware retrieval strategy robustly preserves strict revisit consistency even under complex occlusions.

\paragraph{Memory Injection Mechanism.} We evaluate our adaptive memory injection module in Tab.~\ref{tab:re10k_ablation} (bottom) and Fig.~\ref{fig:ab_adaptive}. All variants are trained on Re10K using the same fine-tuning settings as our final model. 
(1) ``w/o mem.'': the baseline (Wan-CamCtrl~\cite{wan2025wan}) fails to maintain scene consistency during revisits. 
(2) ``w/o align.'': this variant directly conditions the diffusion model via frame-wise concatenation with retrieved historical frames, yielding degraded visual quality and camera control. We attribute this to the difficulty of implicitly learning complex 3D correspondences solely through LoRA fine-tuning. 
(3) ``w/ fz. NVS'': this variant uses a pre-trained frozen LVSM~\cite{jin2024lvsm} to align historical frames with the target view before injection. While revisit consistency improves, extrapolation errors from the frozen NVS module interfere with camera motion, causing navigation failures (e.g., the camera fails to enter the bathroom), as shown in Fig.~\ref{fig:ab_adaptive}. 
(4) ``w/ ft. NVS'': In contrast, using adaptive NVS module ensures both revisit consistency and accurate camera control by jointly fine-tuning the NVS module with the diffusion backbone.

\section{Conclusion}
In this paper, we present \textbf{I3DM}, an implicit 3D-aware memory mechanism designed to ensure long-term scene consistency in video generation. To address limitations of existing explicit geometry-based and implicit multi-view-based memory mechanisms, we propose an implicit 3D-aware memory retrieval strategy to achieve robust, occlusion-aware historical context retrieval without explicit geometry modeling. We further introduce an adaptive 3D-aligned memory injection module that aligns retrieved historical frames with the target view and adaptively guides the generation process. Extensive experiments show that I3DM achieves superior revisit consistency, visual fidelity, and camera control, providing insights for memory mechanisms in future world models.

\let\vspace\sourcevspace

\bibliography{main}

\begin{thebibliography}{50}
\providecommand{\natexlab}[1]{#1}

\bibitem[{Bai et~al.(2025)Bai, Xia, Fu, Wang, Mu, Cao, Liu, Hu, Bai, Wan
  et~al.}]{bai2025recammaster}
Bai, J.; Xia, M.; Fu, X.; Wang, X.; Mu, L.; Cao, J.; Liu, Z.; Hu, H.; Bai, X.;
  Wan, P.; et~al. 2025.
\newblock Recammaster: Camera-controlled generative rendering from a single
  video.
\newblock In \emph{Proceedings of the IEEE/CVF International Conference on
  Computer Vision}, 14834--14844.

\bibitem[{Brooks et~al.(2024)Brooks, Peebles, Holmes, DePue, Guo, Jing,
  Schnurr, Taylor, Luhman, Luhman et~al.}]{brooks2024video}
Brooks, T.; Peebles, B.; Holmes, C.; DePue, W.; Guo, Y.; Jing, L.; Schnurr, D.;
  Taylor, J.; Luhman, T.; Luhman, E.; et~al. 2024.
\newblock Video generation models as world simulators.
\newblock \emph{OpenAI Blog}, 1(8): 1.

\bibitem[{Chen et~al.(2025)Chen, Chen, Xiu, Geiger, and Chen}]{chen2025ttt3r}
Chen, X.; Chen, Y.; Xiu, Y.; Geiger, A.; and Chen, A. 2025.
\newblock Ttt3r: 3d reconstruction as test-time training.
\newblock \emph{arXiv preprint arXiv:2509.26645}.

\bibitem[{Chen et~al.(2024)Chen, Xu, Zheng, Zhuang, Pollefeys, Geiger, Cham,
  and Cai}]{chen2024mvsplat}
Chen, Y.; Xu, H.; Zheng, C.; Zhuang, B.; Pollefeys, M.; Geiger, A.; Cham,
  T.-J.; and Cai, J. 2024.
\newblock Mvsplat: Efficient 3d gaussian splatting from sparse multi-view
  images.
\newblock In \emph{European conference on computer vision}, 370--386. Springer.

\bibitem[{He et~al.(2024)He, Xu, Guo, Wetzstein, Dai, Li, and
  Yang}]{he2024cameractrl}
He, H.; Xu, Y.; Guo, Y.; Wetzstein, G.; Dai, B.; Li, H.; and Yang, C. 2024.
\newblock Cameractrl: Enabling camera control for text-to-video generation.
\newblock \emph{arXiv preprint arXiv:2404.02101}.

\bibitem[{Heusel et~al.(2017)Heusel, Ramsauer, Unterthiner, Nessler, and
  Hochreiter}]{heusel2017fid}
Heusel, M.; Ramsauer, H.; Unterthiner, T.; Nessler, B.; and Hochreiter, S.
  2017.
\newblock Gans trained by a two time-scale update rule converge to a local nash
  equilibrium.
\newblock \emph{Advances in neural information processing systems}, 30.

\bibitem[{Huang et~al.(2025{\natexlab{a}})Huang, Hu, Han, Shi, Tian, He, and
  Jiang}]{huang2025memoryForcing}
Huang, J.; Hu, X.; Han, B.; Shi, S.; Tian, Z.; He, T.; and Jiang, L.
  2025{\natexlab{a}}.
\newblock Memory forcing: Spatio-temporal memory for consistent scene
  generation on minecraft.
\newblock \emph{arXiv preprint arXiv:2510.03198}.

\bibitem[{Huang et~al.(2025{\natexlab{b}})Huang, Zheng, Wang, Liu, Wang, Wu,
  Jiang, Li, Lau, Zuo et~al.}]{huang2025voyager}
Huang, T.; Zheng, W.; Wang, T.; Liu, Y.; Wang, Z.; Wu, J.; Jiang, J.; Li, H.;
  Lau, R.; Zuo, W.; et~al. 2025{\natexlab{b}}.
\newblock Voyager: Long-range and world-consistent video diffusion for
  explorable 3d scene generation.
\newblock \emph{ACM Transactions on Graphics (TOG)}, 44(6): 1--15.

\bibitem[{Huang et~al.(2024)Huang, He, Yu, Zhang, Si, Jiang, Zhang, Wu, Jin,
  Chanpaisit, Wang, Chen, Wang, Lin, Qiao, and Liu}]{huang2023vbench}
Huang, Z.; He, Y.; Yu, J.; Zhang, F.; Si, C.; Jiang, Y.; Zhang, Y.; Wu, T.;
  Jin, Q.; Chanpaisit, N.; Wang, Y.; Chen, X.; Wang, L.; Lin, D.; Qiao, Y.; and
  Liu, Z. 2024.
\newblock {VBench}: Comprehensive Benchmark Suite for Video Generative Models.
\newblock In \emph{Proceedings of the IEEE/CVF Conference on Computer Vision
  and Pattern Recognition}.

\bibitem[{Jia et~al.(2026)Jia, Sun, You, Wong, Zou, Yan, Wu, and
  Jiang}]{jia2026efficient-lvsm}
Jia, X.; Sun, Y.; You, J.; Wong, S.; Zou, Z.; Yan, J.; Wu, Z.; and Jiang, Y.-G.
  2026.
\newblock Efficient-LVSM: Faster, Cheaper, and Better Large View Synthesis
  Model via Decoupled Co-Refinement Attention.
\newblock \emph{arXiv preprint arXiv:2602.06478}.

\bibitem[{Jiang et~al.(2025)Jiang, Mao, Xu, Lu, Ren, Jin, Xu, Yu, Pang, Zhao
  et~al.}]{jiang2025anysplat}
Jiang, L.; Mao, Y.; Xu, L.; Lu, T.; Ren, K.; Jin, Y.; Xu, X.; Yu, M.; Pang, J.;
  Zhao, F.; et~al. 2025.
\newblock Anysplat: Feed-forward 3d gaussian splatting from unconstrained
  views.
\newblock \emph{ACM Transactions on Graphics (TOG)}, 44(6): 1--16.

\bibitem[{Jin et~al.(2024)Jin, Jiang, Tan, Zhang, Bi, Zhang, Luan, Snavely, and
  Xu}]{jin2024lvsm}
Jin, H.; Jiang, H.; Tan, H.; Zhang, K.; Bi, S.; Zhang, T.; Luan, F.; Snavely,
  N.; and Xu, Z. 2024.
\newblock Lvsm: A large view synthesis model with minimal 3d inductive bias.
\newblock \emph{arXiv preprint arXiv:2410.17242}.

\bibitem[{Kendall and Gal(2017)}]{kendall2017uncertainties}
Kendall, A.; and Gal, Y. 2017.
\newblock What uncertainties do we need in bayesian deep learning for computer
  vision?
\newblock \emph{Advances in neural information processing systems}, 30.

\bibitem[{Knapitsch et~al.(2017)Knapitsch, Park, Zhou, and
  Koltun}]{Knapitsch2017tnt}
Knapitsch, A.; Park, J.; Zhou, Q.-Y.; and Koltun, V. 2017.
\newblock Tanks and Temples: Benchmarking Large-Scale Scene Reconstruction.
\newblock \emph{ACM Transactions on Graphics}, 36(4).

\bibitem[{Kong et~al.(2025)Kong, Yang, Zheng, and Wang}]{kong2025worldwarp}
Kong, H.; Yang, X.; Zheng, X.; and Wang, X. 2025.
\newblock WorldWarp: Propagating 3D Geometry with Asynchronous Video Diffusion.
\newblock \emph{arXiv preprint arXiv:2512.19678}.

\bibitem[{Kong et~al.(2024)Kong, Tian, Zhang, Min, Dai, Zhou, Xiong, Li, Wu,
  Zhang et~al.}]{kong2024hunyuanvideo}
Kong, W.; Tian, Q.; Zhang, Z.; Min, R.; Dai, Z.; Zhou, J.; Xiong, J.; Li, X.;
  Wu, B.; Zhang, J.; et~al. 2024.
\newblock Hunyuanvideo: A systematic framework for large video generative
  models.
\newblock \emph{arXiv preprint arXiv:2412.03603}.

\bibitem[{Li et~al.(2025)Li, Torr, Vedaldi, and Jakab}]{li2025vmem}
Li, R.; Torr, P.; Vedaldi, A.; and Jakab, T. 2025.
\newblock VMem: Consistent Interactive Video Scene Generation with
  Surfel-Indexed View Memory.
\newblock \emph{arXiv preprint arXiv:2506.18903}.

\bibitem[{Lipman et~al.(2022)Lipman, Chen, Ben-Hamu, Nickel, and
  Le}]{lipman2022flow}
Lipman, Y.; Chen, R.~T.; Ben-Hamu, H.; Nickel, M.; and Le, M. 2022.
\newblock Flow matching for generative modeling.
\newblock \emph{arXiv preprint arXiv:2210.02747}.

\bibitem[{Mao et~al.(2025)Mao, Lin, Li, Li, Peng, He, Pang, Chi, Qiao, and
  Zhang}]{mao2025yume}
Mao, X.; Lin, S.; Li, Z.; Li, C.; Peng, W.; He, T.; Pang, J.; Chi, M.; Qiao,
  Y.; and Zhang, K. 2025.
\newblock Yume: An interactive world generation model.
\newblock \emph{arXiv preprint arXiv:2507.17744}.

\bibitem[{Parker-Holder et~al.(2024)Parker-Holder, Ball, Bruce, Dasagi,
  Holsheimer, Kaplanis, Moufarek, Scully, Shar, Shi et~al.}]{parker2024genie}
Parker-Holder, J.; Ball, P.; Bruce, J.; Dasagi, V.; Holsheimer, K.; Kaplanis,
  C.; Moufarek, A.; Scully, G.; Shar, J.; Shi, J.; et~al. 2024.
\newblock Genie 2: A large-scale foundation world model.
\newblock \emph{URL: https://deepmind.
  google/discover/blog/genie-2-a-large-scale-foundation-world-model}.

\bibitem[{Peebles and Xie(2023)}]{peebles2023scalable}
Peebles, W.; and Xie, S. 2023.
\newblock Scalable diffusion models with transformers.
\newblock In \emph{Proceedings of the IEEE/CVF international conference on
  computer vision}, 4195--4205.

\bibitem[{Plucker(1865)}]{plucker1865xvii}
Plucker, J. 1865.
\newblock Xvii. on a new geometry of space.
\newblock \emph{Philosophical Transactions of the Royal Society of London},
  (155): 725--791.

\bibitem[{Qin et~al.(2024)Qin, Shi, Yu, Wang, Zhou, Li, Yin, Liu, Sheng, Shao
  et~al.}]{qin2024worldsimbench}
Qin, Y.; Shi, Z.; Yu, J.; Wang, X.; Zhou, E.; Li, L.; Yin, Z.; Liu, X.; Sheng,
  L.; Shao, J.; et~al. 2024.
\newblock Worldsimbench: Towards video generation models as world simulators.
\newblock \emph{arXiv preprint arXiv:2410.18072}.

\bibitem[{Ren et~al.(2025)Ren, Shen, Huang, Ling, Lu, Nimier-David, M{\"u}ller,
  Keller, Fidler, and Gao}]{ren2025gen3c}
Ren, X.; Shen, T.; Huang, J.; Ling, H.; Lu, Y.; Nimier-David, M.; M{\"u}ller,
  T.; Keller, A.; Fidler, S.; and Gao, J. 2025.
\newblock Gen3c: 3d-informed world-consistent video generation with precise
  camera control.
\newblock In \emph{Proceedings of the Computer Vision and Pattern Recognition
  Conference}, 6121--6132.

\bibitem[{Sajjadi et~al.(2022)Sajjadi, Meyer, Pot, Bergmann, Greff, Radwan,
  Vora, Lu{\v{c}}i{\'c}, Duckworth, Dosovitskiy et~al.}]{sajjadi2022srt}
Sajjadi, M.~S.; Meyer, H.; Pot, E.; Bergmann, U.; Greff, K.; Radwan, N.; Vora,
  S.; Lu{\v{c}}i{\'c}, M.; Duckworth, D.; Dosovitskiy, A.; et~al. 2022.
\newblock Scene representation transformer: Geometry-free novel view synthesis
  through set-latent scene representations.
\newblock In \emph{Proceedings of the IEEE/CVF conference on computer vision
  and pattern recognition}, 6229--6238.

\bibitem[{Sargent et~al.(2024)Sargent, Li, Shah, Herrmann, Yu, Zhang, Chan,
  Lagun, Fei-Fei, Sun et~al.}]{sargent2024zeronvs}
Sargent, K.; Li, Z.; Shah, T.; Herrmann, C.; Yu, H.-X.; Zhang, Y.; Chan, E.~R.;
  Lagun, D.; Fei-Fei, L.; Sun, D.; et~al. 2024.
\newblock Zeronvs: Zero-shot 360-degree view synthesis from a single image.
\newblock In \emph{Proceedings of the IEEE/CVF Conference on Computer Vision
  and Pattern Recognition}, 9420--9429.

\bibitem[{Sun et~al.(2025)Sun, Zhang, Wang, Wu, Wang, Wang, Wang, Zhang, Wang,
  and Guo}]{sun2025worldplay}
Sun, W.; Zhang, H.; Wang, H.; Wu, J.; Wang, Z.; Wang, Z.; Wang, Y.; Zhang, J.;
  Wang, T.; and Guo, C. 2025.
\newblock Worldplay: Towards long-term geometric consistency for real-time
  interactive world modeling.
\newblock \emph{arXiv preprint arXiv:2512.14614}.

\bibitem[{Szymanowicz et~al.(2026)Szymanowicz, Chen, Wang, Rupprecht, and
  Vedaldi}]{szymanowicz2026lagernvs}
Szymanowicz, S.; Chen, M.; Wang, J.; Rupprecht, C.; and Vedaldi, A. 2026.
\newblock LagerNVS: Latent Geometry for Fully Neural Real-time Novel View
  Synthesis.
\newblock In \emph{Proceedings of the IEEE/CVF Conference on Computer Vision
  and Pattern Recognition}, 15443--15453.

\bibitem[{Team(2025)}]{qwen2.5-VL}
Team, Q. 2025.
\newblock Qwen2.5-VL.

\bibitem[{Team et~al.(2026)Team, Gao, Wang, Zeng, Zhu, Cheng, Li, Wang, Xu, Ma,
  Chen, Liu, Cheng, Yao, Zhu, Meng, Zheng, Bai, Chen, Shen, Yu, Zhu, Shen, and
  Ouyang}]{lingbot-world}
Team, R.; Gao, Z.; Wang, Q.; Zeng, Y.; Zhu, J.; Cheng, K.~L.; Li, Y.; Wang, H.;
  Xu, Y.; Ma, S.; Chen, Y.; Liu, J.; Cheng, Y.; Yao, Y.; Zhu, J.; Meng, Y.;
  Zheng, K.; Bai, Q.; Chen, J.; Shen, Z.; Yu, Y.; Zhu, X.; Shen, Y.; and
  Ouyang, H. 2026.
\newblock Advancing Open-source World Models.
\newblock \emph{arXiv preprint arXiv:2601.20540}.

\bibitem[{Unterthiner et~al.(2018)Unterthiner, Van~Steenkiste, Kurach,
  Marinier, Michalski, and Gelly}]{unterthiner2018fvd}
Unterthiner, T.; Van~Steenkiste, S.; Kurach, K.; Marinier, R.; Michalski, M.;
  and Gelly, S. 2018.
\newblock Towards accurate generative models of video: A new metric \&
  challenges.
\newblock \emph{arXiv preprint arXiv:1812.01717}.

\bibitem[{Wan et~al.(2025)Wan, Wang, Ai, Wen, Mao, Xie, Chen, Yu, Zhao, Yang
  et~al.}]{wan2025wan}
Wan, T.; Wang, A.; Ai, B.; Wen, B.; Mao, C.; Xie, C.-W.; Chen, D.; Yu, F.;
  Zhao, H.; Yang, J.; et~al. 2025.
\newblock Wan: Open and advanced large-scale video generative models.
\newblock \emph{arXiv preprint arXiv:2503.20314}.

\bibitem[{Wang et~al.(2025{\natexlab{a}})Wang, Chen, Karaev, Vedaldi,
  Rupprecht, and Novotny}]{wang2025vggt}
Wang, J.; Chen, M.; Karaev, N.; Vedaldi, A.; Rupprecht, C.; and Novotny, D.
  2025{\natexlab{a}}.
\newblock Vggt: Visual geometry grounded transformer.
\newblock In \emph{Proceedings of the Computer Vision and Pattern Recognition
  Conference}, 5294--5306.

\bibitem[{Wang et~al.(2025{\natexlab{b}})Wang, Zhang, Holynski, Efros, and
  Kanazawa}]{wang2025cut3r}
Wang, Q.; Zhang, Y.; Holynski, A.; Efros, A.~A.; and Kanazawa, A.
  2025{\natexlab{b}}.
\newblock Continuous 3d perception model with persistent state.
\newblock In \emph{Proceedings of the Computer Vision and Pattern Recognition
  Conference}, 10510--10522.

\bibitem[{Wang et~al.(2025{\natexlab{c}})Wang, Xu, Dai, Xiang, Deng, Tong, and
  Yang}]{wang2025moge}
Wang, R.; Xu, S.; Dai, C.; Xiang, J.; Deng, Y.; Tong, X.; and Yang, J.
  2025{\natexlab{c}}.
\newblock Moge: Unlocking accurate monocular geometry estimation for
  open-domain images with optimal training supervision.
\newblock In \emph{Proceedings of the Computer Vision and Pattern Recognition
  Conference}, 5261--5271.

\bibitem[{Wang et~al.(2024{\natexlab{a}})Wang, Leroy, Cabon, Chidlovskii, and
  Revaud}]{wang2024dust3r}
Wang, S.; Leroy, V.; Cabon, Y.; Chidlovskii, B.; and Revaud, J.
  2024{\natexlab{a}}.
\newblock Dust3r: Geometric 3d vision made easy.
\newblock In \emph{Proceedings of the IEEE/CVF Conference on Computer Vision
  and Pattern Recognition}, 20697--20709.

\bibitem[{Wang et~al.(2025{\natexlab{d}})Wang, Zhou, Zhu, Chang, Zhou, Li,
  Chen, Pang, Shen, and He}]{wang2025pi}
Wang, Y.; Zhou, J.; Zhu, H.; Chang, W.; Zhou, Y.; Li, Z.; Chen, J.; Pang, J.;
  Shen, C.; and He, T. 2025{\natexlab{d}}.
\newblock pi3: Permutation-Equivariant Visual Geometry Learning.
\newblock \emph{arXiv preprint arXiv:2507.13347}.

\bibitem[{Wang et~al.(2004)Wang, Bovik, Sheikh, and Simoncelli}]{wang2004ssim}
Wang, Z.; Bovik, A.~C.; Sheikh, H.~R.; and Simoncelli, E.~P. 2004.
\newblock Image quality assessment: from error visibility to structural
  similarity.
\newblock \emph{IEEE transactions on image processing}, 13(4): 600--612.

\bibitem[{Wang et~al.(2026)Wang, Liu, Li, Huang, Xu, Kang, An, Wang, Jiang, Wei
  et~al.}]{wang2026matrix}
Wang, Z.; Liu, Z.; Li, J.; Huang, K.; Xu, B.; Kang, F.; An, M.; Wang, P.;
  Jiang, B.; Wei, Y.; et~al. 2026.
\newblock Matrix-game 3.0: Real-time and streaming interactive world model with
  long-horizon memory.
\newblock \emph{arXiv preprint arXiv:2604.08995}.

\bibitem[{Wang et~al.(2024{\natexlab{b}})Wang, Yuan, Wang, Li, Chen, Xia, Luo,
  and Shan}]{wang2024motionctrl}
Wang, Z.; Yuan, Z.; Wang, X.; Li, Y.; Chen, T.; Xia, M.; Luo, P.; and Shan, Y.
  2024{\natexlab{b}}.
\newblock Motionctrl: A unified and flexible motion controller for video
  generation.
\newblock In \emph{ACM SIGGRAPH 2024 Conference Papers}, 1--11.

\bibitem[{Wu et~al.(2026)Wu, He, Cheng, Yang, Zhang, Kang, Cai, Wei, Guo, Li
  et~al.}]{wu2026infinite}
Wu, R.; He, X.; Cheng, M.; Yang, T.; Zhang, Y.; Kang, Z.; Cai, X.; Wei, X.;
  Guo, C.; Li, C.; et~al. 2026.
\newblock Infinite-World: Scaling Interactive World Models to 1000-Frame
  Horizons via Pose-Free Hierarchical Memory.
\newblock \emph{arXiv preprint arXiv:2602.02393}.

\bibitem[{Wu et~al.(2025)Wu, Yang, Po, Xu, Liu, Lin, and
  Wetzstein}]{wu2025spaMem}
Wu, T.; Yang, S.; Po, R.; Xu, Y.; Liu, Z.; Lin, D.; and Wetzstein, G. 2025.
\newblock Video World Models with Long-term Spatial Memory.
\newblock \emph{arXiv preprint arXiv:2506.05284}.

\bibitem[{Xiao et~al.(2025)Xiao, Lan, Zhou, Ouyang, Yang, Zeng, and
  Pan}]{xiao2025worldmem}
Xiao, Z.; Lan, Y.; Zhou, Y.; Ouyang, W.; Yang, S.; Zeng, Y.; and Pan, X. 2025.
\newblock Worldmem: Long-term consistent world simulation with memory.
\newblock \emph{arXiv preprint arXiv:2504.12369}.

\bibitem[{Yang et~al.(2026)Yang, Fan, Shi, Peng, Wang, and
  Zhang}]{yang2026neoverse}
Yang, Y.; Fan, L.; Shi, Z.; Peng, J.; Wang, F.; and Zhang, Z. 2026.
\newblock NeoVerse: Enhancing 4D World Model with in-the-wild Monocular Videos.
\newblock \emph{arXiv preprint arXiv:2601.00393}.

\bibitem[{Yu et~al.(2025)Yu, Bai, Qin, Liu, Wang, Wan, Zhang, and
  Liu}]{yu2025cam}
Yu, J.; Bai, J.; Qin, Y.; Liu, Q.; Wang, X.; Wan, P.; Zhang, D.; and Liu, X.
  2025.
\newblock Context as memory: Scene-consistent interactive long video generation
  with memory retrieval.
\newblock In \emph{Proceedings of the SIGGRAPH Asia 2025 Conference Papers},
  1--11.

\bibitem[{Yu et~al.(2024)Yu, Xing, Yuan, Hu, Li, Huang, Gao, Wong, Shan, and
  Tian}]{yu2024viewcrafter}
Yu, W.; Xing, J.; Yuan, L.; Hu, W.; Li, X.; Huang, Z.; Gao, X.; Wong, T.-T.;
  Shan, Y.; and Tian, Y. 2024.
\newblock Viewcrafter: Taming video diffusion models for high-fidelity novel
  view synthesis.
\newblock \emph{arXiv preprint arXiv:2409.02048}.

\bibitem[{Zhang et~al.(2018)Zhang, Isola, Efros, Shechtman, and
  Wang}]{zhang2018lpips}
Zhang, R.; Isola, P.; Efros, A.~A.; Shechtman, E.; and Wang, O. 2018.
\newblock The unreasonable effectiveness of deep features as a perceptual
  metric.
\newblock In \emph{Proceedings of the IEEE conference on computer vision and
  pattern recognition}, 586--595.

\bibitem[{Zhao et~al.(2026)Zhao, Wei, Liu, Zhang, Xu, and Lu}]{zhao2026spatia}
Zhao, J.; Wei, F.; Liu, Z.; Zhang, H.; Xu, C.; and Lu, Y. 2026.
\newblock Spatia: Video generation with updatable spatial memory.
\newblock In \emph{Proceedings of the IEEE/CVF Conference on Computer Vision
  and Pattern Recognition}, 4245--4257.

\bibitem[{Zhou et~al.(2025)Zhou, Gao, Voleti, Vasishta, Yao, Boss, Torr,
  Rupprecht, and Jampani}]{zhou2025seva}
Zhou, J.; Gao, H.; Voleti, V.; Vasishta, A.; Yao, C.-H.; Boss, M.; Torr, P.;
  Rupprecht, C.; and Jampani, V. 2025.
\newblock Stable virtual camera: Generative view synthesis with diffusion
  models.
\newblock \emph{arXiv preprint arXiv:2503.14489}.

\bibitem[{Zhou et~al.(2018)Zhou, Tucker, Flynn, Fyffe, and
  Snavely}]{zhou2018stereo}
Zhou, T.; Tucker, R.; Flynn, J.; Fyffe, G.; and Snavely, N. 2018.
\newblock Stereo magnification: Learning view synthesis using multiplane
  images.
\newblock \emph{arXiv preprint arXiv:1805.09817}.

\end{thebibliography}

\end{document}

\maketitle

% Manual source spacing is disabled to comply with the AAAI format.
\let\sourcevspace\vspace
\renewcommand{\vspace}[1]{}

This supplementary material is organized as follows:
\begin{itemize}
    \item[$\bullet$] Sec.~\ref{sec:limit}: \textbf{Discussion and Limitations.}
    \item[$\bullet$] Sec.~\ref{sec:m_d}: \textbf{Details of 3D-aware Memory Retrieval.}
    \item[$\bullet$] Sec.~\ref{sec:ada_examp}: 
    \textbf{Details of Adaptive Memory Injection.}
    \item[$\bullet$] Sec.~\ref{sec:impl}: \textbf{Implementation Details.}
    \item[$\bullet$] Sec.~\ref{sec:ab_s}: \textbf{Additional Ablation Studies.}
    \item[$\bullet$] Sec.~\ref{sec:vis}: \textbf{Additional Comparisons \& Visual Results.}
\end{itemize}
We highly recommend \textbf{watching our supplementary video} for a clearer visualization of the experimental results.

\section{Discussion and Limitations}
\label{sec:limit} Despite the effectiveness of our memory retrieval and injection mechanisms, certain limitations are discussed below.

\paragraph{Inference Time Analysis.}
We report the average per-frame inference time as the video length increases in Tab.~\ref{tab:inference_breakdown}. The retrieval time grows nearly linearly with video length but remains manageable, accounting for only 15.5\% of total inference time even at 989 frames. This overhead is comparable to WorldMem~\cite{xiao2025worldmem}, which reports a 10--20\% retrieval overhead at 1K frames. Since our focus is accurate retrieval in complex environments, we leave acceleration through spatial pruning or more efficient architectures such as Efficient-LVSM~\cite{jia2026efficient-lvsm} for future improvement.
\begin{table}[H]
    \centering
    \caption{Per-frame runtime (sec) as generated video length increases. \textbf{Bank size} denotes the number of frames in the memory bank for retrieval before generating the current clip.}
    \label{tab:inference_breakdown}
    \resizebox{\linewidth}{!}{%
    \begin{tabular}{ccccc}
        \hline
        \textbf{Frames} & \textbf{Bank size} & \textbf{Retrieval time}
        & \textbf{Total time} & \textbf{Ratio} \\
        \hline
        77  & 4   & 0.00 & 1.24 & 0.0\%  \\
        229 & 44  & 0.04 & 1.32 & 3.1\%  \\
        457 & 104 & 0.10 & 1.36 & 7.3\%  \\
        609 & 144 & 0.14 & 1.39 & 9.9\%  \\
        837 & 204 & 0.19 & 1.47 & 13.2\% \\
        989 & 244 & 0.23 & 1.50 & 15.5\% \\
        \hline
    \end{tabular}
    }
\end{table}

\begin{figure}
  \centering
\includegraphics[width=1.0\linewidth]{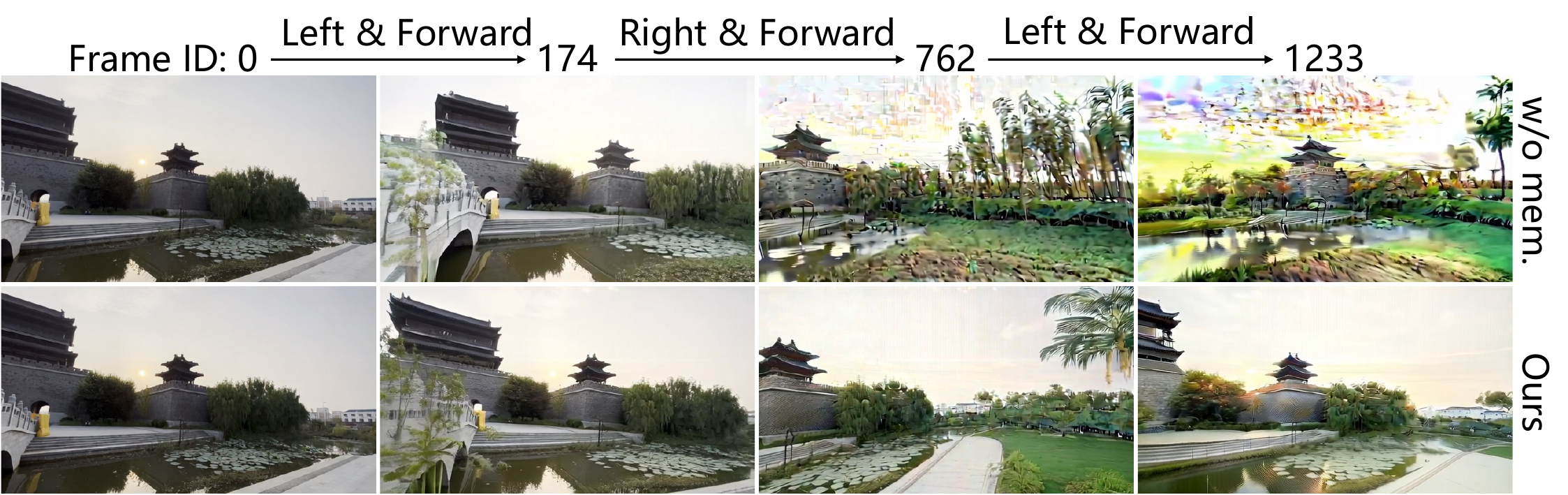}
  \caption{Our method alleviates temporal drifting by memory mechanism.}
  \label{fig:drift}
  % \tightfigspace
\end{figure}

\paragraph{Drifting and Error Accumulation.}
Temporal drifting caused by inherent error accumulation for long video generation remains a challenge, which mainly stems from the train-test gap of the base diffusion model. We report a result in Fig.~\ref{fig:drift}, which compares our method with a variant without memory on 1233-frame generation. Although our memory mechanism can reduce error accumulation by retrieving relatively clean early frames to guide later generation, our method does not focus on solving this problem. There is a line of work (as known as ``Forcing'' methods)~\cite{huang2025self_forcing, liu2025rolling, yang2025longlive} exclusively targeting on this problem, and combining our method with these techniques could further mitigate this issue.
% \beforefloatspace
% \vspace{-1mm}

\begin{figure}
  \centering
\includegraphics[width=1.0\linewidth]{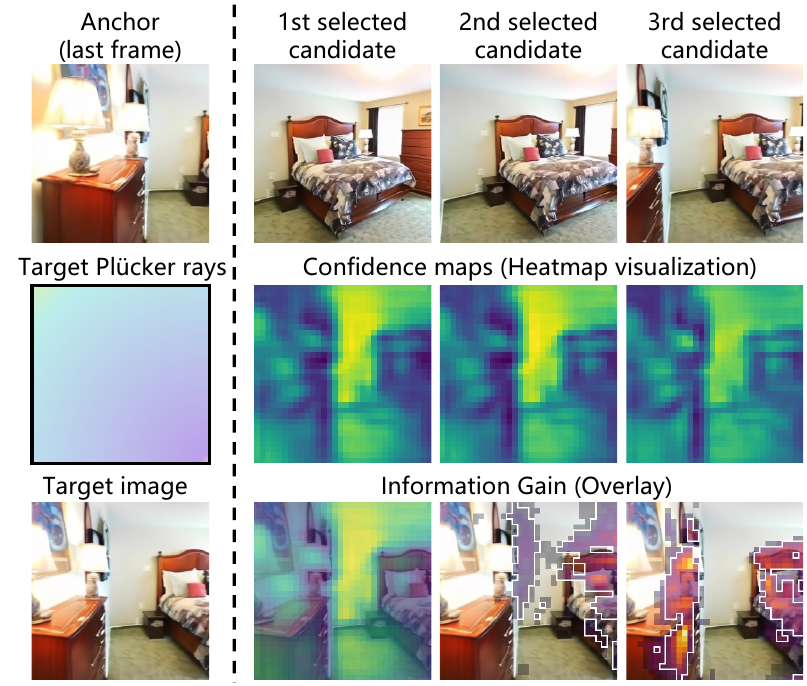}
  \caption{Visualization of confidence maps and greedy maximum-coverage selection. Given the anchor frame and target Plücker rays, our network predicts a spatial confidence map for each candidate (middle row). The selected candidates provide complementary marginal gains that collectively cover the target view (bottom row).}
  \label{fig:vis}
  % \tightfigspace
\end{figure}

\section{Details of 3D-aware Memory Retrieval}
\label{sec:m_d}

\paragraph{Visualization of Confidence Maps.}
Figure~\ref{fig:vis} visualizes the retrieval process for a target view. For each candidate, our network takes the candidate frame, the last frame as the anchor, and the target Plücker rays as inputs, and predicts a target-aligned spatial confidence map indicating how reliably the candidate covers each target-view region (middle row). Based on these confidence maps, the greedy maximum-coverage algorithm sequentially selects three candidates by maximizing the marginal information gain at each step, promoting complementary target-view coverage while reducing redundancy (bottom row). The target image is shown only for reference and is not used during retrieval.

\paragraph{Maximum Coverage Selection Algorithm.}

After obtaining the set of spatial confidence maps for each frame in the memory bank (excluding the last frame) using our proposed scoring CNN, we perform a \textit{greedy maximum coverage selection} to retrieve relevant historical frames that maximally cover the scene of the target query views. The algorithm is summarized in Alg.~\ref{alg:greedy_selection}.

\begin{algorithm}[H]
\caption{Multi-View Maximum Coverage Selection For One Target View}
\label{alg:greedy_selection}
\footnotesize
\begin{algorithmic}[1]
    \renewcommand{\algorithmicrequire}{\textbf{Input:}}
    \renewcommand{\algorithmicensure}{\textbf{Output:}}
    \newcommand{\algdescription}[1]{%
        \Statex \parbox[t]{\linewidth}{\raggedright #1}}
    \newcommand{\alginlinecomment}[1]{%
        \unskip\hfill{\small$\triangleright$~#1}}
    
    \Require
    \algdescription{$i_{\text{last}}$: index of the last frame}
    \algdescription{$N$: number of conditioning frames}
    \algdescription{$\mathcal{V} = \{(\mathbf{I}_i, \mathbf{P}_i)\}_{i=1}^{i_{\text{last}}}$: memory bank of historical frames and Plücker rays.}
    \algdescription{$\mathcal{M} = \{\mathbf{m}_i\}_{i=1}^{i_{\text{last}}-1}$: candidate confidence maps, with $\mathbf{m}_i \in \mathbb{R}^{\frac{H}{p} \times \frac{W}{p}}$}
    
    \Ensure 
    \algdescription{$\mathcal{C}$: indices of selected frames}
    \Statex \rule{\linewidth}{0.4pt}
    
    \State $\mathcal{C} \gets \emptyset$ \alginlinecomment{Initialize the selected indices}
    \State $K \gets \min(N-1, |\mathcal{V}|-1)$ \alginlinecomment{Number of frames to retrieve}
    \State $\mathcal{I} \gets \{1, 2, \dots, i_{\text{last}}-1\}$ \alginlinecomment{Indices in the memory bank}
    \State $\mathbf{m}^\text{g} \gets \mathbf{0}^{\frac{H}{p} \times \frac{W}{p}}$ \alginlinecomment{Initialize global confidence}
    
    \For{$step = 1$ \textbf{to} $K$}
        \State $g^{*} \gets -\infty$
        \State $i^{*} \gets -1$
        
        \For{\textbf{each} $i \in \mathcal{I}$}
            \State $\tilde{\mathbf{m}}^{\text{g}} \gets \max(\mathbf{m}^{\text{g}}, \mathbf{m}_i)$ \alginlinecomment{Element-wise maximum}
            \State $g \gets \sum (\tilde{\mathbf{m}}^{\text{g}} - \mathbf{m}^{\text{g}})$ \alginlinecomment{Calculate the information gain}
            
            \If{$g > g^{*}$}
                \State $g^{*} \gets g$
                \State $i^{*} \gets i$
            \EndIf
        \EndFor
        
        \If{$i^{*} \neq -1$}
            \State $\mathcal{C} \gets \mathcal{C} \cup \{i^{*}\}$ \alginlinecomment{Include the best candidate}
            \State $\mathcal{I} \gets \mathcal{I} \setminus \{i^{*}\}$ \alginlinecomment{Remove from candidate pool}
            \State $\mathbf{m}^{\text{g}} \gets \max(\mathbf{m}^{\text{g}}, \mathbf{m}_{i^{*}})$ \alginlinecomment{Update global confidence}
        \EndIf
    \EndFor
    
    \State $\mathcal{C} \gets \mathcal{C} \cup \{i_{\text{last}}\}$ \alginlinecomment{Include the last frame}
    \State \Return $\mathcal{C}$
\end{algorithmic}
\end{algorithm}

\begin{figure*}[t]
  \centering
  \includegraphics[width=1.0\textwidth]{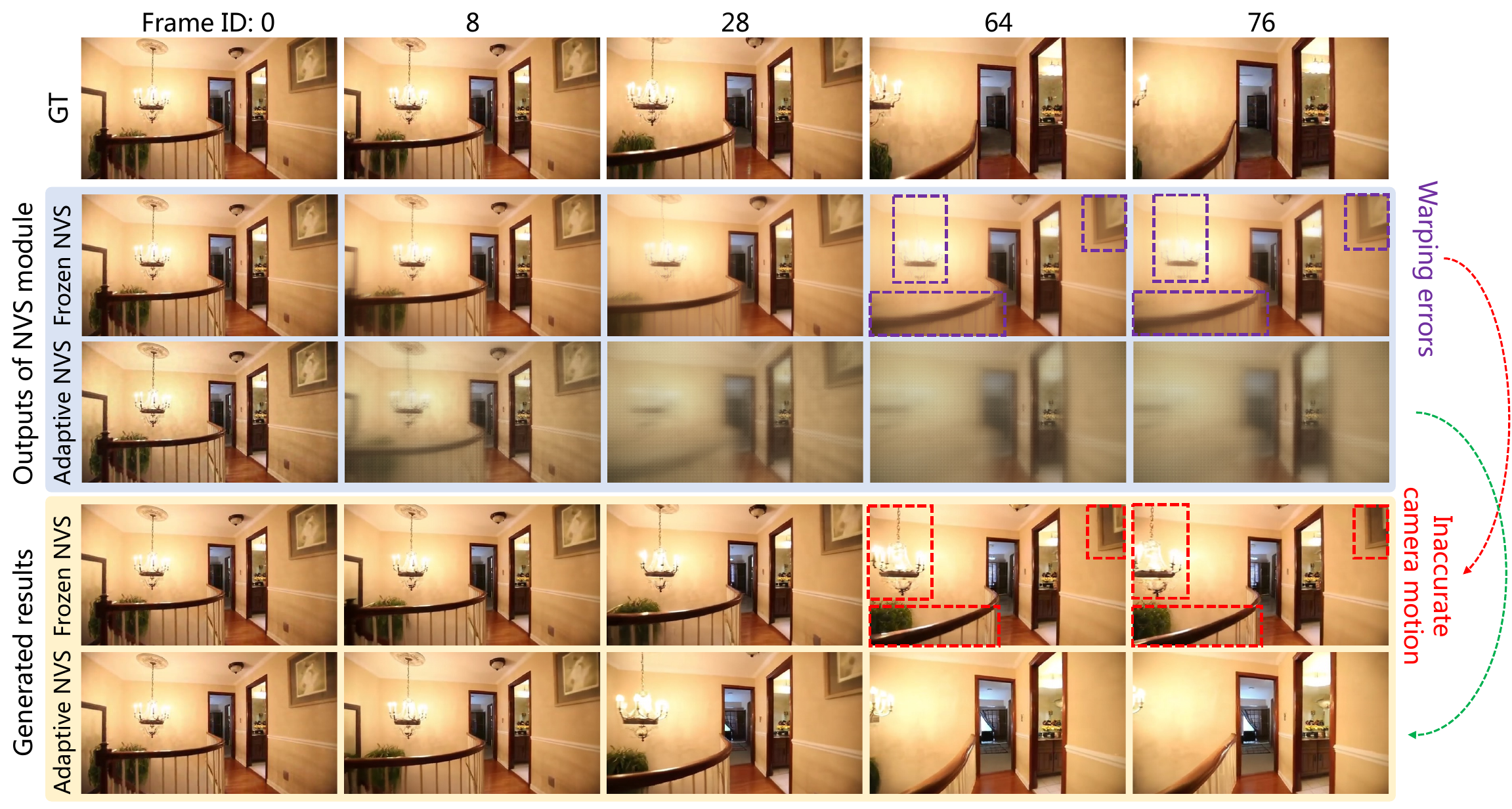}
  \caption{Visual comparison of the frozen and adaptive NVS modules. The purple dashed boxes indicate warping errors introduced by the frozen NVS module, which causes inaccurate camera motion in the generated results (highlighted by red dashed boxes). The adaptive NVS module suppresses these unreliable warped regions, allowing the model to fall back on its internal generative priors to produce convincing results with accurate camera motion.
  }
  \label{fig:supp_adaptive}
\vspace{-0.5em}
\end{figure*}

\section{Details of Adaptive Memory Injection}
\label{sec:ada_examp}
We utilize the pre-trained LVSM~\cite{jin2024lvsm} as our NVS module to warp the retrieved historical frames to the target view, providing spatially aligned guidance to condition the diffusion model. However, the frozen NVS module produces unreliable warping outputs in extrapolated regions (Fig.~\ref{fig:supp_adaptive}, second row), providing misleading guidance that interferes with the diffusion model and causes inaccurate camera motion in the generated results (Fig.~\ref{fig:supp_adaptive}, fourth row). In contrast, jointly fine-tuning the NVS module with the diffusion model enables it to produce clear guidance in reliably aligned regions while suppressing unreliable extrapolated content (Fig.~\ref{fig:supp_adaptive}, third row), thereby adaptively guiding the diffusion model to generate plausible results with accurate camera motion (Fig~\ref{fig:supp_adaptive}, fifth row).

\section{Implementation Details} 
\label{sec:impl}

\paragraph{Training of Memory Retrieval Module.} We utilize the decoder-only variant of the pre-trained LVSM~\cite{jin2024lvsm} as the feature extractor. Input images are resized to $256\times256$, and features are extracted from the $6^{\text{th}}$ Transformer layer. The scoring CNN has 3 layers with output channels of 256, 64, and 1, respectively. We freeze the entire LVSM model and solely train the scoring CNN on the RealEstate10K~\cite{zhou2018stereo} training set using a learning rate of $5 \times 10^{-5}$ and a total batch size of 64. We train it for 16k steps, which takes approximately 10 hours on 4 NVIDIA RTX 4090 GPUs. 

\paragraph{Training of Video Generation Module.} Our video generation model is based on Wan2.1-Fun-V1.1-1.3B-Control-Camera~\cite{wan2025wan, he2024cameractrl}. We generate video clips at a resolution of $640\times352$ with 77 frames, conditioned on 4 frames randomly sampled from the same scene. The Adaptive NVS Module is initialized from the pre-trained LVSM. Because the original pretrained LVSM does not support 4 input views and arbitrary resolutions, we solely fine-tune the LVSM decoder-only variant for 6k steps using its original configuration, except for the number of input views and frame resolution. After that, we train our video generation model by jointly fine-tuning the Transformer blocks of the Adaptive NVS Module, the reference convolution layer, and the LoRA layers injected into the DiT blocks for 11k steps. The rank of the LoRA layers is set to 1024. Training employs the AdamW optimizer with a learning rate of $1 \times 10^{-4}$ and a total batch size of 4. The training process takes approximately 1.8 days on 4 NVIDIA H200 GPUs. 

\paragraph{Camera Control Metrics.}
Following existing work~\cite{li2025vmem, kong2025worldwarp, bai2025recammaster, zhang2025ucpe, he2024cameractrl}, we express estimated camera poses relative to the first frame and normalize the translation by the furthest frame. After extracting poses of generated views using Pi3~\cite{wang2025pi}, we calculate the rotation error ($R_{err}^\circ$) and the translation error ($T_{err}$) against the ground truth as follows:
\begin{align}
    R_{{err}}^\circ & = \arccos \left( 0.5 \left( \text{tr}(\mathbf{R}_{\text{gen}}\mathbf{R}_{\text{gt}}^T) - 1 \right) \right), \\
    T_{{err}} & = \| \mathbf{T}_{\text{gt}} - \mathbf{T}_{\text{gen}} \|_2,
\end{align}
where $\mathbf{R}_{\text{gen}}$ and $\mathbf{T}_{\text{gen}}$ denote the rotation matrix and translation vector of generated views, and $\mathbf{R}_{\text{gt}}$ and $\mathbf{T}_{\text{gt}}$ denote their ground truth counterparts. $\text{tr}$ denotes the trace of a matrix.

\begin{table*}[t]
    \centering
    \caption{Ablation studies of different feature layers for memory retrieval on the Re10K dataset. Time costs are averaged per video clip and reported in seconds (s).}
    \label{tab:ab_layer}
    
    \setlength{\aboverulesep}{0pt}
    \setlength{\belowrulesep}{0pt}
    
    \renewcommand{\arraystretch}{1.1}
    
    \newcolumntype{S}{>{\hsize=0.9\hsize\centering\arraybackslash}X}
    \newcolumntype{Z}{>{\hsize=1.25\hsize\centering\arraybackslash}X}
    
    \begin{tabularx}{\textwidth}{l | S S | S S S | Z Z}
        \toprule
        \multirow{2}{*}{Layer} & \multicolumn{2}{c|}{\textbf{Visual Quality}} & \multicolumn{3}{c|}{\textbf{Revisit Consistency}} & \multicolumn{2}{c}{\textbf{Per-clip Time Cost}} \\
        
        \cmidrule(lr){2-3} \cmidrule(lr){4-6} \cmidrule(lr){7-8}
        
        & FID $\downarrow$ & FVD $\downarrow$ & PSNR $\uparrow$ & SSIM $\uparrow$ & LPIPS $\downarrow$ & Retrieval $\downarrow$ & Generation $\downarrow$ \\
        \midrule
        
        L0  & 18.655 & 148.764 & 17.183 & 0.634 & 0.2764 & 0.1853 & 56.7241 \\
        L4  & 18.098 & 133.384 & 24.609 & 0.823 & 0.0786 & 1.3481 & 58.3389 \\
        L6  & 17.553 & 131.657 & 24.732 & 0.828 & 0.0756 & 1.9739 & 59.0871 \\
        L12 & 17.656 & 134.054 & 24.771 & 0.828 & 0.0761 & 3.8737 & 61.4180 \\
        L24 & 17.599 & 128.981 & 24.823 & 0.830 & 0.0742 & 7.6167 & 64.9117 \\
        \bottomrule
    \end{tabularx}
\end{table*}

\section{Additional Ablation Studies}
\label{sec:ab_s}

\paragraph{Choice of Feature Layer.}
We evaluate the impact of using different intermediate feature layers from the pre-trained LVSM~\cite{jin2024lvsm} to predict the spatial uncertainty map for historical frame retrieval. Tab.~\ref{tab:ab_layer} reports the visual quality and revisit consistency evaluated over the entire generated videos (6 consecutive clips), and the average per-clip (77 frames) time cost. Here, L0 (raw tokens before Transformer blocks) yields the poorest revisit consistency, indicating that features without spatial correspondence learning fail to provide effective 3D priors for retrieval. Features extracted from deeper layers yield more accurate uncertainty maps, improving revisit consistency. However, this performance gain comes at the cost of linearly increasing retrieval times. To strike a good balance, we adopt the $6^{\text{th}}$ layer in our default implementation. At this layer, the retrieval overhead accounts for $3.3\%$ of the total per-clip generation time, which is acceptable.

\begin{figure*}[!p]
  \centering
  \includegraphics[width=0.98\textwidth]{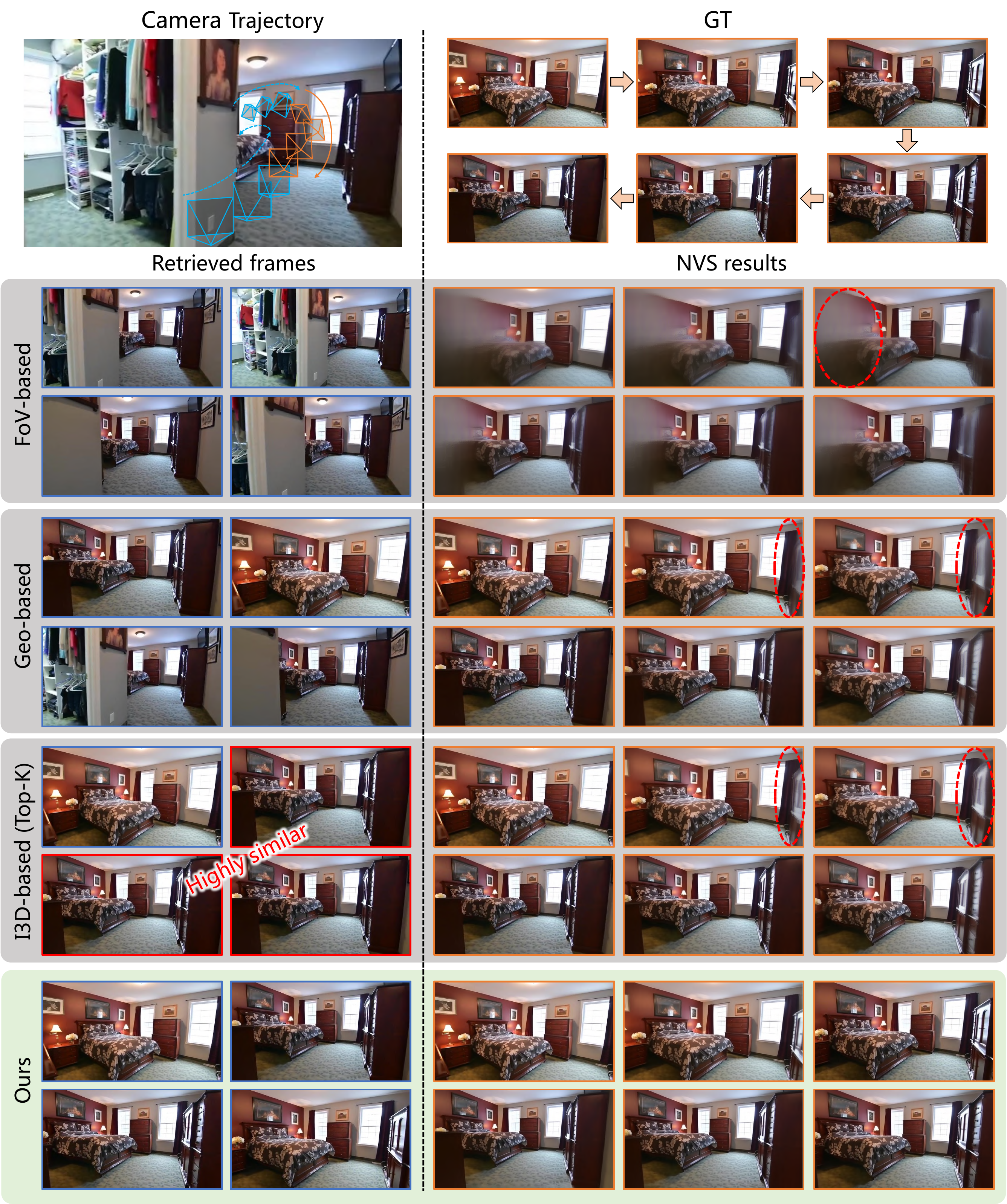}
  \caption{Qualitative comparison of different memory retrieval strategies. (Top) The camera trajectory illustrates the memory bank initialization (blue) and the target sequence to be generated (orange), alongside the ground truth (GT) frames. (Bottom) For each strategy, the left panel shows the four retrieved historical frames, while the right panel displays the corresponding target views synthesized by the pre-trained NVS model. Red dashed circles highlight artifacts caused by suboptimal retrieval.}
  \label{fig:supp_retrieve}
\end{figure*}

\paragraph{Qualitative Analysis of Retrieval Strategies.} 
To further validate our implicit 3D-aware memory retrieval strategy, we conduct a comparative experiment, as illustrated in Fig.~\ref{fig:supp_retrieve}. 
Specifically, we initialize the memory bank using ground-truth images from the first four video clips (blue dashed trajectories in Fig.~\ref{fig:supp_retrieve}) and aim to generate the fifth clip (orange dashed trajectory). 
Using the target camera poses of the fifth clip, we retrieve relevant historical frames from the memory bank to evaluate four different selection strategies: FoV-based, geometry-based, I3D-based (Top-K), and our proposed approach. 
Note that the memory bank contains the same ground-truth frames across all strategies for a fair comparison.
% Specifically, we initialize the memory bank using ground-truth images from the first four video clips (indicated by the blue dashed trajectories in Fig.~\ref{fig:supp_adaptive}) and aim to generate the fifth clip (the orange dashed trajectory). Using the target camera poses from the fifth clip, we retrieve relevant historical frames from the memory bank (which composes of the same frames using ground truth images for a fair comparison) to evaluate the four different strategies.
These retrieved frames (shown on the left of Fig.~\ref{fig:supp_retrieve}) are then fed into the pre-trained NVS model~\cite{jin2024lvsm} to synthesize the target views (shown on the right). 

As demonstrated by the results, the FoV-based strategy retrieves frames where the target region is occluded by the wardrobe, leading to severe information loss and blurry NVS results. While the geometry-based approach mitigates this occlusion issue, it still exhibits noticeable artifacts. The I3D-based (Top-K) method retrieves highly similar frames with redundant information overlap, resulting in blurry synthesis in regions lacking spatial reference. In contrast, our full implicit 3D-aware retrieval strategy successfully selects unoccluded frames while maximizing complementary scene coverage for the target view, ultimately yielding the highest NVS quality.

\begin{figure}
  \centering
  \includegraphics[width=\linewidth]{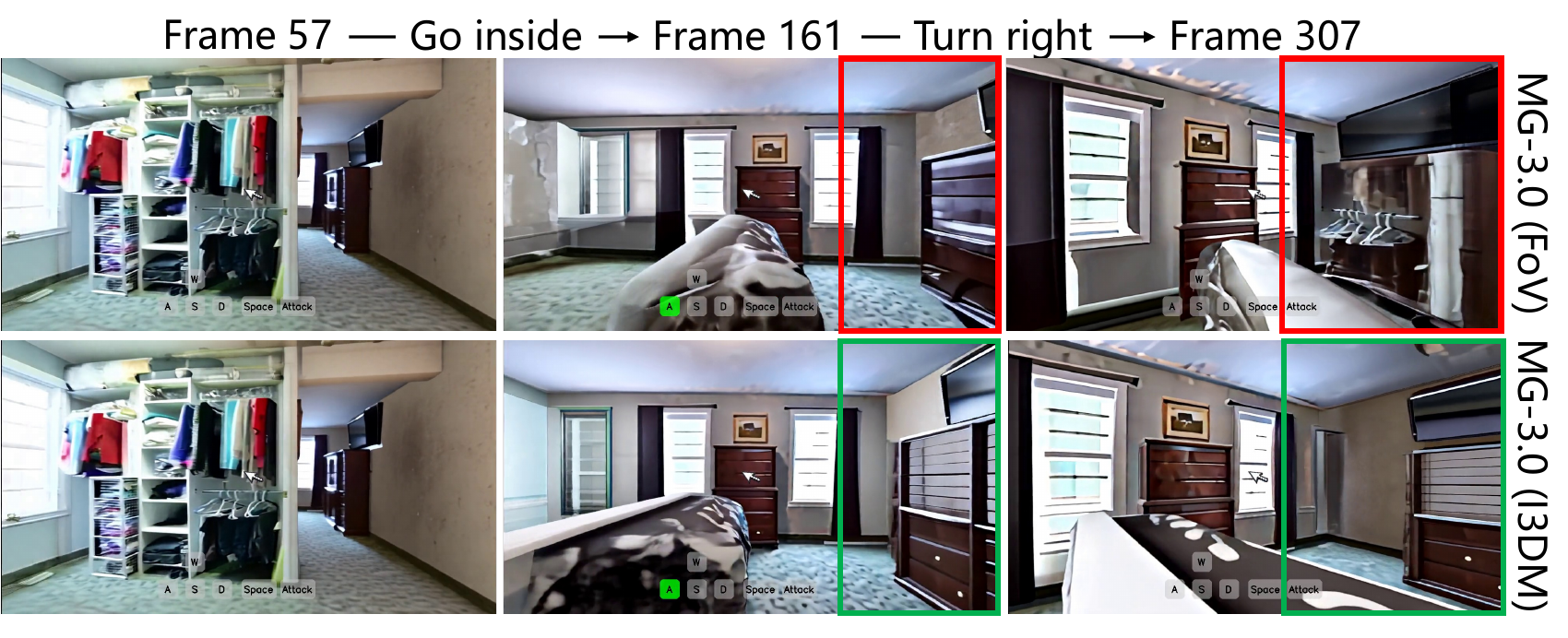}
  \caption{Replacing the FoV-based retrieval of Matrix-Game-3.0~\cite{wang2026matrix} with our 3D-aware retrieval strategy improves revisit consistency while retaining its original frame-concatenation injection.}
  \label{fig:matrix}
\end{figure}

\paragraph{Generality of Our Retrieval Strategy.}
Our implicit 3D-aware memory retrieval strategy can be integrated into other memory-based video generation frameworks. For example, Matrix-Game-3.0~\cite{wang2026matrix} uses FoV overlap for memory retrieval and frame concatenation for memory injection. Replacing only its FoV-based retrieval with ours, while retaining the original injection mechanism, improves revisit consistency, as shown in Fig.~\ref{fig:matrix}.

% \begin{figure*}[!t]
%   \centering
%   \includegraphics[width=0.98\textwidth]{fig/supp_retri_frames_v1_compressed.pdf}
%   \caption{Qualitative comparison of different memory retrieval strategies. (Top) The camera trajectory illustrates the memory bank initialization (blue) and the target sequence to be generated (orange), alongside the ground truth (GT) frames. (Bottom) For each strategy, the left panel shows the four retrieved historical frames, while the right panel displays the corresponding target views synthesized by the pre-trained NVS model. Red dashed circles highlight artifacts caused by suboptimal retrieval.}
%   % The FoV-based strategy retrieves heavily occluded frames (e.g., blocked by the wardrobe). Geo-based and I3D-based (Top-K) methods fail to provide sufficient complementary coverage, resulting in noticeable edge artifacts. In contrast, our approach successfully retrieves unoccluded, spatially relevant frames, achieving best NVS results.}
%   \label{fig:supp_retrieve}
% % \vspace{-0.5em}
% \end{figure*}

% \section{Additional Comparisons \& Visual Results}
% \label{sec:vis}

\section{Additional Comparisons \& Visual Results}
\label{sec:vis}

\begin{figure*}[!p]
  \centering
  \includegraphics[width=1.0\textwidth]{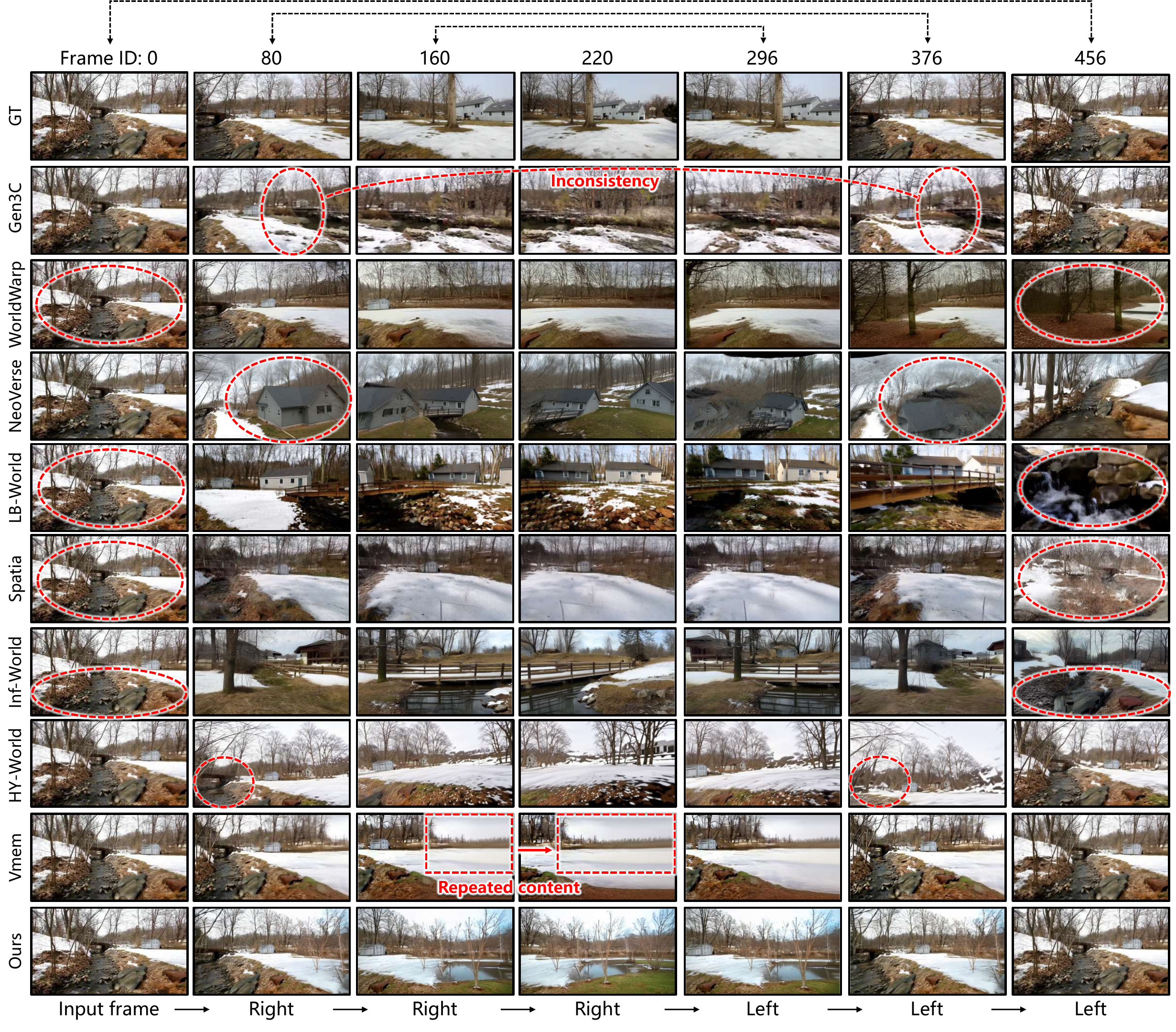}
  \caption{Qualitative comparison on an outdoor RealEstate10K scene. Black dashed arrows link corresponding frames that should remain consistent; red dashed circles highlight visual inconsistencies, red dashed arrows indicate inaccurate camera motion, and red dashed boxes denote repetitive generated content.}
  \label{fig:supp_comp_outdoor}
\end{figure*}

\begin{figure*}[!p]
  \centering
  \includegraphics[width=1.0\textwidth]{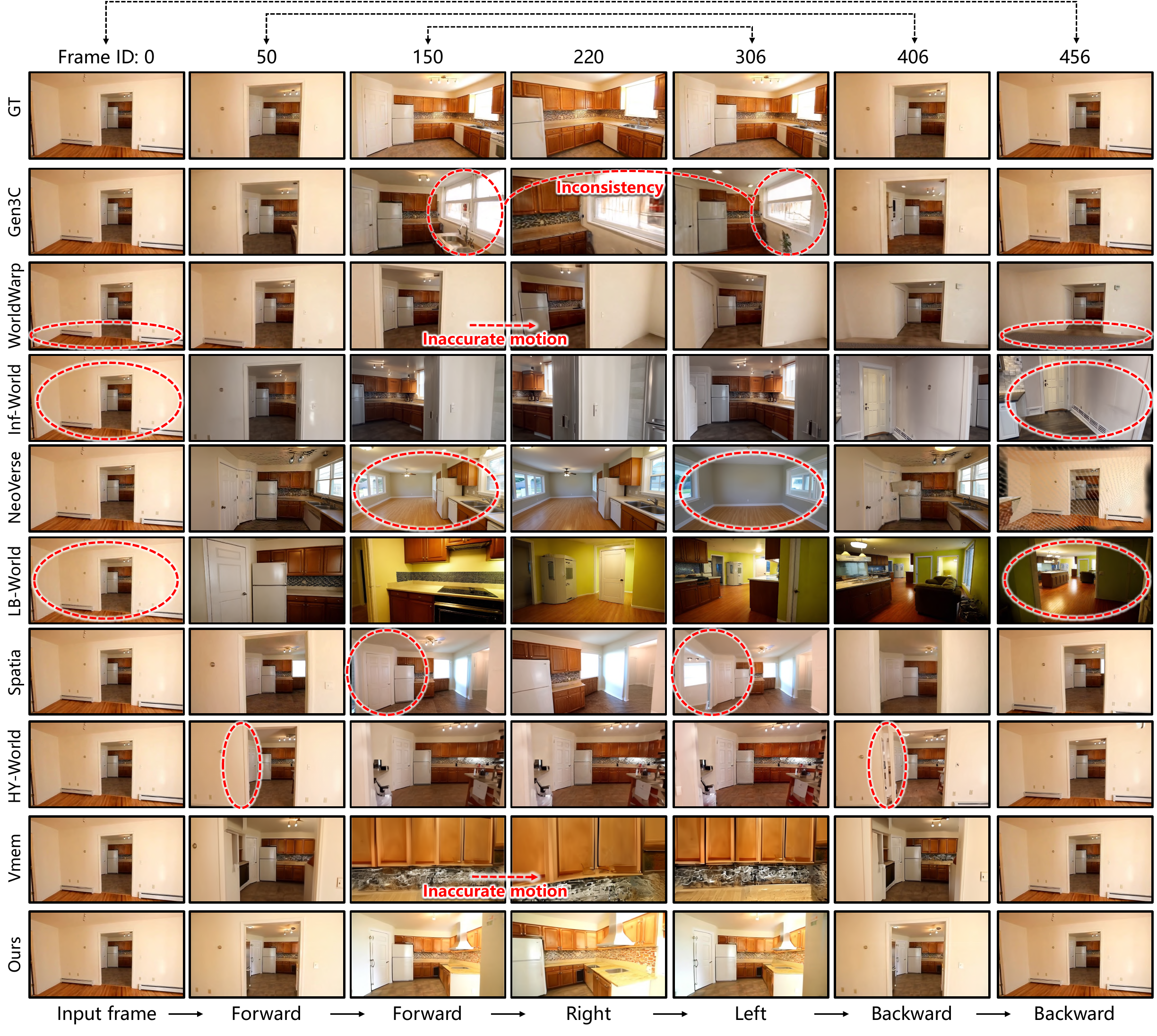}
  \caption{Qualitative comparison on an indoor RealEstate10K scene. Black dashed arrows link corresponding frames that should remain consistent; red dashed circles highlight visual inconsistencies.}
  \label{fig:supp_comp_indoor}
\end{figure*}

\paragraph{Qualitative Comparison with Baselines.} We provide additional qualitative comparisons in Figs.~\ref{fig:supp_comp_outdoor} and~\ref{fig:supp_comp_indoor}. As highlighted, most works~\cite{kong2025worldwarp, ren2025gen3c, wu2026infinite, yang2026neoverse, lingbot-world, zhao2025spatia, sun2025worldplay} fail to maintain strict consistency during revisits. Meanwhile, Vmem~\cite{li2025vmem} tends to produce repetitive content and exhibits inaccurate camera motion. In contrast, our method generates plausible scene content during exploration and maintains strict consistency upon revisiting, all while accurately adhering to the target camera trajectory.

\begin{figure*}[!p]
  \centering
  \includegraphics[width=1.0\textwidth]{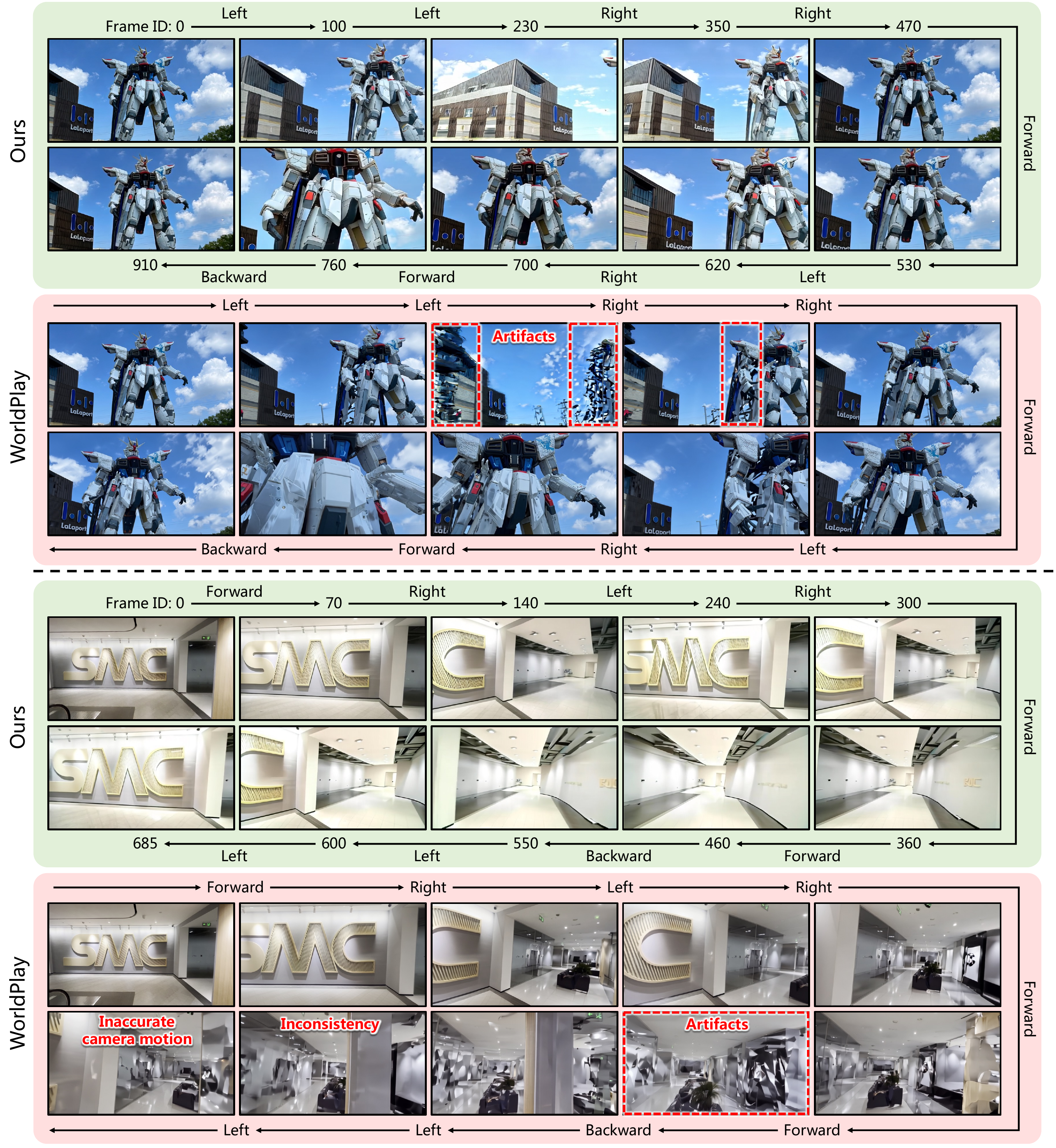}
  \caption{Qualitative comparison on the out-of-distribution scenes. Red dashed boxes denote the generated artifacts.}
  \label{fig:supp_ood}
\end{figure*}

\paragraph{OOD Generalization.} We conduct further generalization experiments on out-of-distribution scenes, as shown in Fig.~\ref{fig:supp_ood}. We capture two real-world scenes and synthesize long camera trajectories (over 910 and 680 frames, respectively) starting from the reference photo, comparing our method against WorldPlay~\cite{sun2025worldplay}. As demonstrated in Fig.~\ref{fig:supp_ood}, our method generates plausible novel content during exploration and maintains good consistency during revisits. Although WorldPlay exhibits less color shifting due to its large-scale training data, it still suffers from severe artifacts during exploration, as highlighted by the red dashed boxes in Fig.~\ref{fig:supp_ood}.

% \begin{figure}[t]
%   \centering
%   \includegraphics[width=\linewidth]{fig/matrix_v2.pdf}
%   \caption{Replacing the FoV-based retrieval of Matrix-Game-3.0~\cite{wang2026matrix} with our 3D-aware retrieval strategy improves revisit consistency while retaining its original frame-concatenation injection.}
%   \label{fig:matrix}
% \end{figure}

\begin{figure*}[!p]
  \centering
  \includegraphics[width=0.84\textwidth]{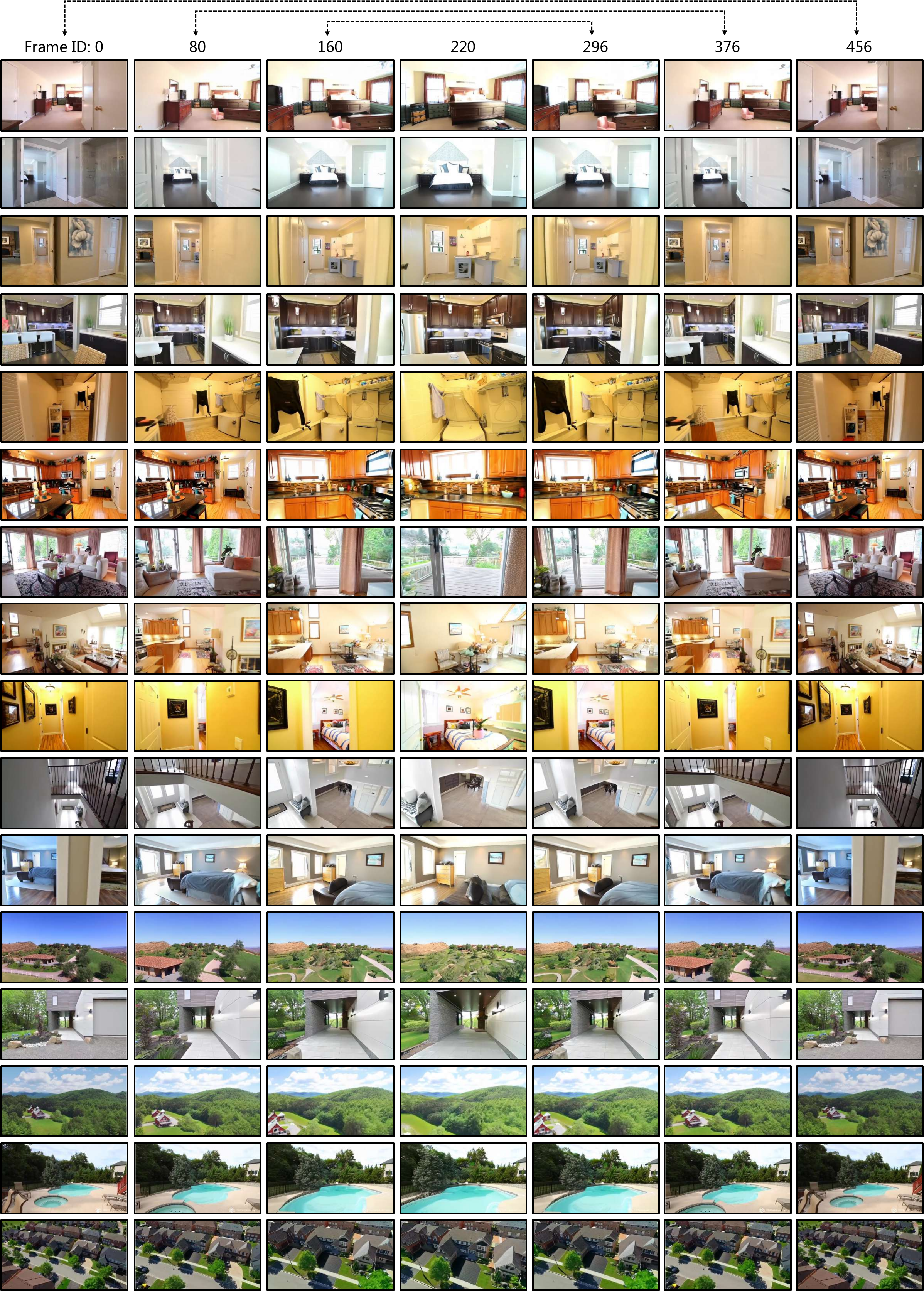}
  \caption{Our visual results on the RealEstate10K dataset. Black dashed arrows link corresponding frames that should remain consistent.}
  \label{fig:supp_ours_re10k}
\end{figure*}

\paragraph{Results on the RealEstate10K Dataset.} We provide additional visual results on the RealEstate10K dataset in Fig.~\ref{fig:supp_ours_re10k}. As demonstrated, our method successfully maintains strict scene consistency while revisiting the same regions.

\clearpage

% \paragraph{Visual Comparison with Other Methods.}

% \paragraph{Additional Visual Results of Re10K Dataset.}

% \paragraph{Additional Visual Results of Out-of-distribution scenes.}
% ---- Bibliography ----
% \let\vspace\sourcevspace
\clearpage
% \onecolumn
% \twocolumn
% \raggedbottom
\bibliography{main}